# Efficient Generative AI Boosts Probabilistic Forecasting of Sudden Stratospheric Warmings

## Author Information

Ningning Tao[1], Fei Xie[1*], Baoxiang Pan[2], Hongyu Wang[1], Han Huang[1], Zhongpu Qiu[1], Ke Gui[4], Jiali Luo[5], Xiaosong Chen[1,3*]

## Affiliations

1. School of Systems Science & Institute of Nonequilibrium Systems, Beijing Normal University, Beijing, China

2. Institute of Atmospheric Physics, Chinese Academy of Sciences, Beijing, China

3. Institute for Advanced Study in Physics and School of Physics, Zhejiang University, Hangzhou, China

4. State Key Laboratory of Severe Weather & Key Laboratory of Atmospheric Chemistry of CMA, Chinese Academy of Meteorological Sciences, Beijing 100081, China

5. College of Atmospheric Sciences, Lanzhou University, Lanzhou 730000, China

## Contributions

F.X., N.T. and X.C. designed the research, conceived the study, carried out the analysis, and prepared the manuscript. N.T. performed the numerical calculations and led the manuscript writing efforts. B.P., H.W., H.H., Z.Q., K.G. and J.L. discussed the results, and assisted in manuscript writing. All authors have read and agreed to the published version of the manuscript.

### Corresponding authors

Correspondence to: Dr. Fei Xie (xiefei@bnu.edu.cn) and Dr. Xiaosong Chen (chenxs@bnu.edu.cn)

## Abstract

Sudden Stratospheric Warmings (SSWs) are key sources of subseasonal predictability and major drivers of extreme weather in winter. Accurate and efficient probabilistic forecasting of these events remains a persistent challenge for Numerical Weather Prediction (NWP) systems due to computational bottlenecks and limitations in physical representation. While data-driven forecasting is rapidly evolving, its application to the complex, three-dimensional dynamics of SSWs remains underexplored. Here, we bridge this gap by developing a Flow Matching-based generative AI model (FM-Cast) for efficient and skillful probabilistic forecasting of the spatiotemporal evolution of stratospheric circulation in winter. Evaluated across 18 major SSW events (1998–2024), FM-Cast successfully forecasts the onset, intensity, and 3D morphology of the polar vortex up to 15 days in advance for most cases. Notably, it achieves long-range probabilistic forecast skill comparable to or exceeding leading operational NWP systems (ECMWF and CMA) while generating a 30-day forecast with 50-member ensemble, in just two minutes on a consumer GPU. Furthermore, using idealized "perfect troposphere" experiments, we uncover distinct predictability regimes: events driven by continuous wave forcing versus those governed by an initial trigger and subsequent stratospheric dynamical memory. This work establishes a computationally efficient paradigm for probabilistic stratospheric forecasting that simultaneously deepens our physical understanding of atmosphere-climate dynamics.

## Introduction

The stratosphere, spanning roughly 10–50 km above the Earth's surface, plays a pivotal yet often underappreciated role in regulating the planet's weather and climate[1]. Beyond housing the ozone layer that shields life from harmful radiation[2], the stratosphere plays a critical role in the dynamical coupling between its own slow, large-scale circulation and the rapidly evolving weather systems below[1]. Variations in stratospheric circulation can propagate downward, altering tropospheric jet streams, storm tracks, and regional weather regimes over timescales of weeks to months[3-7]. These top–down influences make the stratosphere a crucial source of extended-range predictability[7]. A prime example of this influence, and one of the most dramatic events in the winter hemisphere, is the Sudden Stratospheric Warming (SSW)[8]. An SSW involves the complete breakdown of the strong westerly circumpolar vortex; within a few days, it is replaced by weak easterly winds as the polar region experiences a rapid temperature increase of up to 50°C[8]. The impacts of this abrupt reconfiguration are far-reaching, propagating downward to significantly alter tropospheric weather patterns. Specifically, the impact of SSWs on the troposphere manifests through complex dynamical pathways. On the one hand, the anomalous signals in the stratosphere can extend downward, driving a negative phase of the North Atlantic Oscillation (NAO) and favoring the persistence of Greenland Blocking regimes[8-10]. On the other hand, the variations of the stratospheric polar vortex can reflect upward-propagating planetary waves back downward, exerting an indirect dynamic impact on tropospheric weather patterns[12-14]. Collectively, these circulation anomalies facilitate the southward outbreak of Arctic air, leading to an increased risk of extreme cold spells and heavy snowfall across North America and Eurasia[9-14]. Consequently, the accurate forecast of an SSW's onset and evolution is of paramount importance for mitigating the societal and economic risks associated with extreme winter weather.

Significant strides have been made in Numerical Weather Prediction (NWP), driven by improvements in vertical resolution and the parameterization of physical processes like atmospheric waves[8]. Current NWP systems typically provide reliable forecasts for SSW onset at lead times of approximately 10 to 14 days, although this limit varies significantly by event[8,15]. While some highly predictable cases, such as the 2019 SSW, are anticipated up to 18 days in advance[8,16], others remain challenging even at shorter lead times[8,17-20]. However, this event-to-event variability underscores persistent challenges. The accuracy of SSW forecasts remains highly sensitive to uncertainties in both initial conditions and the model's physical representations[15,21-23]. To generate reliable probabilistic guidance, ensemble forecasting is essential, yet this approach creates a major operational bottleneck.

Running dozens of high-resolution model members for multi-week forecasts demands immense computational resources[24-27], motivating the search for new, computationally efficient paradigms.

Deep learning offers a new avenue for efficient weather and climate prediction, with models succeeding in tasks from multi-year ENSO forecasts to medium-range global weather, achieving performance that rivals or surpasses operational NWP systems in standard deterministic metrics[28,29]. For ensemble forecasting, generative AI holds particular promise[30]. In contrast to deterministic AI models, which tend to forecast the mean state of atmospheric evolution and often produce overly smooth or blurry results in long-range forecasts[31], generative AI models sample from a learned data probability distribution. This approach not only overcomes the issue of forecast smoothing but also holds the potential to provide a more reliable quantification of forecast uncertainty, especially in capturing the probability of extreme events[32-40]. In addition, while ensembles can be generated from deterministic models using perturbation methods, these attempts have not addressed the issue of blurring—meaning their members do not represent realistic samples from the weather distribution—and they have not rivaled operational systems[31,32]. Generative AI natively produces realistic and diverse ensembles at a fraction of the computational cost of NWP, positioning it as a powerful tool for probabilistic forecasting[32,40].

While AI weather forecasting has made significant progress, applications to the stratosphere, especially probabilistic forecasts of SSWs, are sparse. Early studies focused on forecasting time-series indices of the polar vortex[41,42] or reconstructing vortex structures without assessing forecast skill[43]. More recently, video prediction models have forecasted the 2D morphology of the vortex at a single pressure level[44]. However, these efforts have not addressed the full spatiotemporal evolution of SSWs across their vertical extent. SSWs, as large-scale atmospheric circulation anomalies representing stratosphere-troposphere coupling and a kind of extreme event, require generative AI for probabilistic forecasting of their onset, intensity, and three-dimensional spatiotemporal evolution.

To bridge this gap, we develop a generative AI model based on Flow Matching, named FM-Cast (Fig. 1), designed for rapid, low-cost, and skillful probabilistic forecasts of the three-dimensional atmospheric fields across the stratosphere (100–10 hPa). We comprehensively evaluate its performance on major SSW events from recent history and, through idealized 'perfect troposphere' experiments, investigate the physical drivers of SSW prediction skill. Our results demonstrate that FM-Cast not only achieves long-range probabilistic forecast skill comparable to that of leading NWP models, but also offers a substantial advantage in computational efficiency, thereby presenting a new and viable paradigm for SSW ensemble forecasting.

## Results

---

To evaluate the forecast skill of FM-Cast across a comprehensive set of Sudden Stratospheric Warming (SSW) events, we selected the period from 1998 to 2024 for our evaluation. This timeframe encompasses 18 major SSW events identified in previous studies[45-49]. To ensure rigorous testing, we partitioned the dataset into three distinct periods using a cross-validation strategy, with a separate model trained independently for each test set (see Methods). The testing periods cover: (1) 1998–2004 (7 SSWs; trained on 1980–1997 and 2005–2024); (2) 2006–2013 (6 SSWs; trained on 1980–2005 and 2014–2024); and (3) 2018–2024 (5 SSWs; trained on 1980–2017). This approach allows us to assess the model's performance on all 18 events, avoiding the selection bias associated with analyzing only specific case studies.

### Probabilistic Forecast Skill and Benchmarking against NWP

We evaluated the forecast skill of FM-Cast using the polar vortex index (defined as the zonal mean zonal wind at 10 hPa and 60°N), alongside the Anomaly Correlation Coefficient (ACC) and Root Mean Square Error (RMSE) metrics (see Methods). The 10 hPa zonal wind serves as the primary indicator for characterizing SSW evolution. Fig. 2 illustrates the ensemble forecast skill of FM-Cast for the polar vortex index across all 18 SSW events. The corresponding ACC and RMSE metrics for the zonal wind fields over the 30°N–89°N domain are presented in Supplementary Fig. S1 and S2, respectively. The results indicate that FM-Cast not only accurately forecasts the temporal evolution of the polar vortex index for most SSW events more than two weeks in advance but also demonstrates high skill in predicting the spatial evolution of the 10 hPa zonal wind field.

Specifically, the model successfully captures diverse dynamical behaviors, including the steep decline of the index and the subsequent wind reversal (e.g., Fig. 2a, b, e, h, j, m, o), as well as complex non-monotonic trends where the vortex briefly strengthens before weakening (e.g., Fig. 2r). For highly predictable events, such as the 2019 SSW[16], the forecast skill for the polar vortex index trend extends beyond 20 days (Fig. 2o). More broadly, FM-Cast exhibits robust forecast capability for the entire stratospheric zonal wind field. The ACC metrics reveal that, even at a 15-day lead time, the ensemble mean ACC remains above 0.5 for the majority of events (Fig. S1 a, b, e, g, h, j, l, m, o, r), indicating that the forecasted patterns are highly correlated with ERA5 reanalysis. Correspondingly, the RMSE plots demonstrate that the 15-day forecast error on the SSW onset day is generally maintained below 20 m/s (Fig. S2). However, for SSW events known to be inherently difficult to forecast, such as the 2009 and 2018 events[17-20], the forecast skill of FM-Cast was correspondingly limited, failing to provide effective guidance beyond a 10-day lead time (k and n of Figs. 2, S1, S2). This alignment with the documented performance drops of NWP models for these specific events suggests that the limitation stems from the intrinsic dynamical complexity of the events rather than a specific deficiency in the FM-Cast model architecture[17-20].

Fig. 3 provides a detailed comparison of the forecast skill for the polar vortex index between FM-Cast and hindcasts from leading NWP models (ECMWF and CMA) at a 15-day lead time for the 18 major SSW events from 1998 to 2024. Results for other lead times are presented in the Supplementary Figs. S4–S6. As a computationally efficient generative AI model, FM-Cast is capable of generating 50 ensemble members with forecast horizons exceeding 30 days within minutes. However, given the limited ensemble size available in the hindcast datasets (11 members for ECMWF and 4 for CMA), we utilized a random subset of 11 FM-Cast ensemble members for the comparisons shown in Fig. 3 to ensure fairness. It is worth noting that the ensemble mean performance of FM-Cast is relatively insensitive to the ensemble size (Fig. S3). Overall, at a 15-day lead time, FM-Cast exhibits a dynamical evolution similar to that of state-of-the-art NWP models, particularly in its ability to predict the deceleration of the vortex. In specific instances (e.g., Fig. 3 a, g, m), FM-Cast achieves a lower absolute errors compared to the NWP models. Specifically, FM-Cast successfully captures the wind reversal (crossing the U=0 m/s threshold) and the subsequent recovery, whereas the NWP models completely fail to capture the breakdown at this lead time, erroneously maintaining strong westerly winds.

Despite accurate performance at longer lead times (e.g., 15 days), FM-Cast exhibits a tendency to overestimate the vortex strength at the timing of the wind reversal at shorter lead times (5 – 10 days) for specific events, such as those in 2003, 2007, 2010, and 2024 (Fig. 2f, i, l, q). Interestingly, even for these cases, FM-Cast maintains a relatively high ACC for the spatial wind field (Fig. S1 f, i, l, q). These events can generally be characterized as "marginal SSWs", where the polar vortex teeters on the edge of collapse or barely crosses the zero-wind threshold before recovering rapidly[50]. Crucially, this difficulty is not unique to the data-driven approach; operational NWP models also struggle significantly with these marginal events[50]. A detailed inspection reveals systematic failures in the NWP models for these cases. For the 2000 event (c of Figs. S4–S6) and 2010 event (l of Figs. S4–S6), both ECMWF and CMA fail to capture the steepness of the decline or the weakening trend across all lead times (10, 7, and 5 days), often missing the reversal entirely; For the 2001 event (e of Figs. S4–S6), CMA consistently fails to predict the vortex weakening at 10, 7, and 5-day lead times; For the 2007 event (i of Figs. S4–S5), both NWP models underestimate the weakening at a 10-day lead time, with ECMWF continuing to underestimate it at 7 days. The shared struggle of both AI and physics-based models in predicting these marginal cases suggests that they represent a distinct class of events with lower intrinsic predictability. The underlying mechanisms governing these threshold-straddling events and the reasons for their low predictability warrant further investigation.

While ECMWF may exhibit higher precision in short-term forecasts (Figs. S4 − S6) — a result expected given that NWP models explicitly solve the governing physical equations—this precision comes at a significant cost. The precise numerical integration of partial differential equations (PDEs), particularly for ensemble forecasting, is computationally prohibitive[24-27]. Consequently, computational bottlenecks[51], combined with the inherent chaotic growth of errors within the PDE systems[52], restrict the extended-range predictability of traditional NWP models. In contrast, as a lightweight generative AI model, FM-Cast leverages data-driven patterns. While it may exhibit lower precision in the short term compared to traditional NWP, its data-driven approach allows it to mitigate, to some extent, the rapid error growth associated with the chaotic nature of the atmosphere and model imperfections, thereby achieving competitive performance at longer lead times. Furthermore, its computational efficiency enables the generation of large-membership ensembles, providing superior probabilistic guidance and robust uncertainty quantification for extreme events such as SSWs.

To quantitatively evaluate the ensemble forecast skill of FM-Cast and compare it with leading NWP models (ECMWF and CMA), we assessed the probability of successful SSW prediction. A forecast is deemed successful if the polar vortex index reverses to easterlies within a specific tolerance window (see Methods for detailed criteria). The ensemble forecast accuracy is defined as the probability of success derived from the ensemble members. However, given the limited ensemble size of the historical hindcast datasets (e.g., only 4 to 11 members for NWP models), the simple frequency counting method results in a coarse probabilistic resolution (e.g., discrete probability steps of ~0.09 for an 11-member ensemble or 0.25 for a 4-member ensemble), which can lead to imprecise probability estimations. To address this limitation and provide a more reliable probability estimation, we adopted the Gaussian PDF-based method following Karpechko et al. (2018)[46]. This approach assumes the ensemble distribution follows a normal distribution defined by the ensemble mean and standard deviation, allowing for a continuous and robust probability assessment even with small sample sizes. Given the variability in predictability across different events, the average ensemble forecast accuracy across the 18 major SSW events serves as a representative metric for the model's overall performance.

Fig. 4 presents the comparison of this average accuracy (with error bars) between FM-Cast and the NWP models using the Gaussian PDF-based method. On average, the probabilistic forecasting skill of FM-Cast exceeds that of the leading NWP models at longer lead times, highlighting the inherent advantage of the data-driven approach in capturing the non-linear dynamics of stratospheric variability. Furthermore, sensitivity analyses confirm that these conclusions are robust to both the choice of probability calculation method and the ensemble size (see Fig. S7).

To further address concerns regarding the model's general performance beyond specific extreme events and to quantify the risk of false alarms, we evaluated the reliability of the probabilistic forecasts. Fig. S8 shows the reliability diagram for SSW predictions across all testing periods, constructed following the methodology of Karpechko (2018)[46] (see Methods). Ideally, for a perfectly calibrated system, the observed frequency of the event should match the forecasted probability (i.e., the curve should lie on the diagonal). As shown in Fig. S8, FM-Cast exhibits a high degree of reliability. Crucially, in the low-probability bins (0–0.2), the observed frequency is correspondingly low, indicating that the model successfully suppresses false alarms during normal stratospheric conditions. This confirms that FM-Cast does not over-forecast SSW occurrences—a critical requirement for operational utility where the event occurrence is not known a priori. At higher probability thresholds, the curve generally follows the diagonal, demonstrating that when the model outputs a high confidence level, an SSW is indeed highly likely to occur.

## Forecasting the Three-Dimensional Stratospheric Evolution

In addition to its skill in forecasting the zonal wind at 10 hPa, FM-Cast demonstrates robust performance for multi-variable fields across multiple vertical levels. To quantify this capability, we selected three representative events—2013, 2018, and 2019—and analyzed the evolution of the Anomaly Correlation Coefficient (ACC) for various variables with forecast lead time. We define the effective forecast horizon as the maximum number of days from the initialization date for which the ACC remains above 0.5. The results indicate that the model's forecast skill is highly correlated with the intrinsic predictability of the event itself. For the highly predictable 2013 and 2019 SSW events[16,53], the effective forecast horizon for most variables at and above 100 hPa reaches 10–15 days (Figs. S9, S11). In contrast, for the 2018 SSW event, which was characterized by low predictability across multiple operational models[18–20], the effective forecast horizon is significantly shorter (Fig. S10).

Fig. 5 showcases an 18-day forecast of the zonal wind field from 70 hPa to 10 hPa generated by FM-Cast, using the 2013 SSW event as an example. The corresponding forecasts for the meridional wind, temperature, geopotential height, and potential vorticity (PV) fields are presented in Supplementary Figs. S12–S15. To quantitatively evaluate forecast performance and examine the spatial distribution of errors, we present the ensemble mean forecasts alongside difference plots (Model minus ERA5) for the wind, temperature, and geopotential height fields. The results demonstrate that FM-Cast accurately forecasts the evolution of various fields at different altitudes, with low error magnitudes confirming its ability to capture the reversal of zonal winds (Fig. 5) and the dramatic warming in the polar region (Fig. S13). Notably, a key advantage of generative AI is their ability to preserve high-frequency details and avoid the "blurring" effect characteristic of deterministic AI models[31,32]. This capability is explicitly illustrated in the PV field (Fig. S15). Unlike other variables where ensemble averaging is used to isolate the signal, for PV, we present a single representative ensemble member. This visualization highlights FM-Cast's ability to maintain sharp gradients and clearly resolve the distinct morphology of the polar vortex splitting, preventing the smoothing of structural details that typically occurs in deterministic AI[31,32].

## Tropospheric Predictability and Stratospheric Memory

As evidenced in Figs. S9 – S11, FM-Cast's forecast skill for the stratosphere significantly exceeds that for the troposphere, consistent with previous findings that the stratosphere possesses higher intrinsic predictability[15]. However, given that tropospheric wave activity is crucial for the generation and development of SSWs[8], a critical question arises: to what extent does the limit of tropospheric predictability constrain the stratospheric forecast?

Fig. S16 and S17 illustrate the ACC evolution for tropospheric 500 hPa geopotential height (Z500) and stratospheric 10 hPa geopotential height (Z10) across 18 SSW events. A distinct divergence in predictability is evident. For most SSW events, the ACC for Z500 drops below 0.5 within 6 – 7 days (Fig. S16). This rapid decline is primarily attributed to the chaotic nature of the troposphere, which is dominated by high-frequency baroclinic instability, leading to rapid error growth that limits the horizon of deterministic prediction. Additionally, limitations in the model's training domain (30 ° N – 89 ° N) and resolution may also contribute to this tropospheric barrier. In contrast, the stratospheric skill (Z10) persists significantly longer, often maintaining high ACC values well after the tropospheric skill has degraded (Fig. S17).

This divergence reflects the fundamental timescale separation between the two layers. The stratosphere exhibits higher dynamical inertia, and major SSW events are often preceded by a "vortex preconditioning" phase that can last for weeks[54]. We hypothesize that once the model captures this preconditioned state from the initial inputs, the subsequent macroscopic evolution of the SSW becomes largely state-dependent and deterministic. Consequently, the model can maintain skill in predicting the stratospheric breakdown (Z10) based on this "system memory," even as its ability to predict specific synoptic tropospheric weather patterns (Z500) diminishes.

Nevertheless, since stratospheric anomalies are dynamically driven by upward-propagating planetary waves from the troposphere, the accumulation of errors in the tropospheric forecast cannot be entirely ignored. In its current form, FM-Cast employs a simplified forecasting scheme for the troposphere. Consequently, accumulated errors in tropospheric variables (e.g., in the amplitude and phase of planetary waves) could eventually decouple the stratosphere from its forcing source, potentially limiting the maximum achievable forecast horizon. To quantify this impact and isolate the influence of tropospheric accuracy on SSW prediction, we designed an idealized experiment termed the "perfect troposphere experiment" In this setup, the tropospheric variables (temperature and geopotential height at 500 hPa and 850 hPa) were constrained to their true values during the autoregressive forecast, while the stratosphere was allowed to evolve freely.

The results of the "perfect troposphere" experiment for the polar vortex index across 18 events are shown in Fig. 6, with additional metrics presented in Supplementary Figs. S18 − S21. The impact of the perfect troposphere varies significantly across events. For some SSWs, correcting the troposphere markedly improved the long-range stratospheric forecast skill. This was particularly evident for the difficult-to-predict events of 2009 and 2018[17-20] (k, n of Figs. 6, S18 vs. k, n of Figs. 2, S17). Taking the 2018 SSW as an example, forcing the model with a perfect troposphere led to a substantial increase in the forecast ACC for all variables from 300 hPa to 10 hPa, with most values exceeding 0.7 (Fig. S20). The effective forecast horizon for all variables at 10 hPa increased from less than 10 days to over 20 days (Fig. S20 vs. Fig. S10), and the forecasted polar vortex index tracked the true index almost perfectly for up to 35 days (blue dotted line in Fig. 6n). Conversely, for other events such as the 2013 SSW, providing perfect tropospheric forcing yielded limited improvement, particularly in the middle and upper stratosphere (m of Figs. 6, S18 vs. m of Figs. 2, S17). Although the perfect troposphere improved the forecast in the lower stratosphere (300 – 100 hPa), the skill enhancement in the middle and upper stratosphere was limited, with the ACC remaining comparatively low (Fig. S19 vs. Fig. S9; Fig. S18m vs. Fig. S18m).

This divergence in model response aligns with the understanding that SSW triggering involves a complex interplay between anomalous tropospheric wave forcing and stratospheric modulation (vortex preconditioning)[54]. Events like those in 2009 and 2018 were characterized by significant, sustained upward EP flux prior to onset[17,18]. These events were likely "forcing-driven," making their forecast skill highly dependent on the accurate representation of the continuous upward wave propagation from the troposphere.

In contrast, the 2013 event highlights a critical distinction between the tropospheric triggering phase and the subsequent stratospheric dynamical evolution. As identified by recent studies, this specific SSW was mechanically "tipped" by a precursor signal—specifically, the strengthening of a North Atlantic ridge approximately 10 days prior to the onset[55]. Notably, our retrospective analysis of the control forecast reveals that FM-Cast accurately captured this crucial tropospheric precursor. As shown in Supplementary Fig. S22, the forecasted geopotential height fields (Z500) during the pre-SSW phase exhibit high fidelity to the ERA5 verification, with minimal error in the key North Atlantic region, particularly during the critical 10-day window preceding the SSW onset. With this initial tropospheric forcing accurately resolved, the polar vortex is effectively driven into a "preconditioned" state[54]. Once this critical regime is reached, the subsequent breakdown dynamics appear to be increasingly governed by the evolution of the preconditioned vortex state, becoming relatively less sensitive to the specific details of continuous tropospheric forcing. This context elucidates why the "perfect troposphere" experiment yielded limited improvement for this specific case. While the "perfect troposphere" setup ensured accurate lower-boundary forcing, the vortex response was already constrained by its strong preconditioning. Therefore, the remaining forecast errors likely stem from limitations in modeling these upper-stratospheric dynamics—potentially pointing to the need for higher vertical resolution or refined upper-level physics—rather than biases in tropospheric forcing.

## Discussion

This study introduces FM-Cast, a generative AI framework for the ensemble forecasting of Sudden Stratospheric Warmings (SSWs) that achieves performance comparable to leading Numerical Weather Prediction (NWP) models at a fraction of the computational cost. On a single consumer-grade GPU, FM-Cast generates a 30-day forecast with 50-member ensemble of key three-dimensional stratospheric fields—including winds, temperature, and geopotential height—in approximately two minutes. This breakthrough in efficiency allows for large-ensemble generation, providing robust uncertainty quantification and avoiding the smoothing artifacts that typically degrade long-range forecasts in deterministic AI models.

Despite its success on major events, FM-Cast, like its NWP counterparts, faces challenges in forecasting "marginal SSWs"—events where the polar vortex teeters on the edge of breakdown. Our comparative analysis shows that systematic failures in predicting the magnitude of deceleration for these events are shared across both data-driven and physics-based systems. This suggests that marginal SSWs represent a distinct class of events with lower intrinsic predictability. The underlying dynamical mechanisms governing these threshold-straddling events remain complex and require further investigation.

Beyond its operational potential, FM-Cast serves as a powerful tool for probing atmospheric dynamics. Our idealized "perfect troposphere" experiments reveal that the predictability of SSWs is governed by distinct dynamical regimes. For events like the 2018 SSW, forecast skill is dominated by continuous tropospheric forcing, where accurate long-range prediction requires sustained input of upward-propagating waves. The profound improvement seen in these experiments unequivocally underscores the critical importance of accurate tropospheric conditions. Currently, our model is constrained by limited tropospheric vertical levels, relatively coarse resolution, and a restricted regional domain, all of which can degrade tropospheric forecast accuracy. Therefore, a primary avenue for future improvement is to significantly strengthen the tropospheric representation—specifically by expanding to a global domain and increasing tropospheric resolution—which will iteratively boost stratospheric predictive capabilities. In contrast, events like the 2013 SSW exhibit a mechanism governed by initial triggering and stratospheric inertia: once the initial tropospheric triggering signal is captured, the subsequent stratospheric evolution is largely determined by the vortex's dynamical memory and preconditioned state. This finding explains why correcting the troposphere during the forecast yields limited skill enhancement for such events and underscores the necessity of developing comprehensive, whole-atmosphere AI models. Future models must also increase vertical resolution within the stratosphere to better resolve stratospheric wave-mean flow interactions and the evolution of these preconditioned states.

It is worth noting that while dynamical NWP models often require post-processing steps, such as bias correction or calibration, to mitigate stratospheric biases and achieve reliability[56-57], FM-Cast achieves well-calibrated probabilistic output intrinsically through its data-driven learning process, without the application of any a posteriori bias correction. However, data-driven models are inherently tethered to the quality of their training data. We acknowledge that the ERA5 training data exhibits a known cold bias in the lower stratosphere during the 2000–2006 period[58]. To assess the impact of this artifact, we conducted a comprehensive sensitivity analysis by retraining the model with the bias-corrected ERA5.1 dataset. The results, detailed in Supplementary Text B and Figs. S23–S28, reveal that while the general structural evolution of the polar vortex is often preserved, noticeable discrepancies in forecast trajectories emerge for specific events. These differences highlight the model's sensitivity to the temporal inconsistencies introduced by mixing reanalysis versions and the limited sample size of extreme stratospheric events available for training. To further bridge the gap with physical consistency and reduce the strict reliance on data homogeneity, future iterations could integrate governing physical equations as soft constraints during the Flow Matching optimization. Additionally, extending the model input to include explicit diagnostics of wave activity (e.g., Eliassen-Palm flux) and expanding the vertical domain to higher stratospheric levels could further enhance skill for both forcing-driven and memory-dominated events. Overall, FM-Cast establishes generative AI as a viable, efficient, and scientifically insightful paradigm for the next generation of Earth system forecasting.

# Methods

## Data Pretreatments

This study utilizes the ERA5 reanalysis dataset[59]. We selected data for zonal wind, meridional wind, temperature, geopotential height, and potential vorticity at five stratospheric pressure levels (100 hPa, 70 hPa, 50 hPa, 10 hPa, and 1 hPa). We also included temperature and geopotential height at four tropospheric levels (850 hPa, 500 hPa, 300 hPa, and 200 hPa). We used the ERA5 dataset covering the period from 1979 to 2024. This period was selected to align with the modern satellite era, ensuring the reliability and consistency of the stratospheric data used for model training and evaluation. As the Northern Hemisphere polar vortex is a winter phenomenon, our dataset comprises data from October, November, December, January, February, March, and April of each year. The temporal resolution is daily (averaged from 6-hourly data), and the spatial resolution is 4° in longitude by 1° in latitude covering the domain from 30°N to 89°N across all longitudes.

To prepare the data for neural network training, we performed several preprocessing steps. Since the lifecycle of the Arctic polar vortex spans across calendar years, we concatenated data from October, November, and December of one year with January, February, March, and April of the following year, designating the winter season by the latter year. This results in a 212-day season for non-leap years and a 213-day season for leap years. To maintain a consistent data structure, we removed April 30th from leap years, ensuring all seasons have a length of 212 days. The final dataset is structured as a five-dimensional tensor with dimensions (45, 212, 33, 60, 90). These dimensions represent the year (45 years from 1980 to 2024), the day of the season, the variables (33 channels combining the five variables at 100–1 hPa and the two variables at 200–850 hPa), latitude, and longitude, respectively. The inclusion of tropospheric variables is intended to capture the influence of tropospheric wave activity on SSWs.

To evaluate the forecast skill of FM-Cast on the 18 major SSW events between 1998 and 2024, and to ensure that each evaluation was performed on strictly unseen data, we partitioned the data into three distinct periods. A separate model was trained independently for each period.

Period 1: Trained on 1980–2017 data (Contains 23 major SSW events), tested on 2018–2024 data (Contains 5 major SSW events).

Period 2: Trained on 1980–2005 and 2014–2024 data (Contains 22 major SSW events), tested on 2006–2013 data (Contains 6 major SSW events).

Period 3: Trained on 1980–1997 and 2005–2024 data (Contains 21 major SSW events), tested on 1998–2004 data (Contains 7 major SSW events).

Given that the input variables have different physical units and numerical ranges, we normalized all variables before training. Normalization eliminates scale discrepancies, preventing the optimization process from being biased towards variables with larger magnitudes, which is crucial for accelerating model convergence and enhancing training stability. We normalized the data by subtracting the spatiotemporal mean and dividing the spatiotemporal standard deviation, which are defined as follows:

$$m_c = \frac{1}{N_y \times 212 \times 60 \times 90} \sum_{y=1}^{N_y} \sum_{d=1}^{212} \sum_{\phi=1}^{60} \sum_{\lambda=1}^{90} X(y, d, c, \phi, \lambda)$$

$$\sigma_c = \sqrt{\frac{1}{N_y \times 212 \times 60 \times 90} \sum_{y=1}^{N_y} \sum_{d=1}^{212} \sum_{\phi=1}^{60} \sum_{\lambda=1}^{90} \left( X(y, d, c, \phi, \lambda) - m_c \right)^2}$$

where $N_y$ is the number of years in the training set, $c$ denotes each variable at each level, and $\phi$ and $\lambda$ represent latitude and longitude, respectively. The spatiotemporal mean and standard deviation were calculated exclusively from the training set for each period. For zonal wind, meridional wind, and potential vorticity, where the sign has a direct physical meaning, we normalized only by dividing by the standard deviation. For temperature and geopotential height, we subtracted the mean before dividing by the standard deviation.

To comprehensively evaluate our model's forecast skill, we used hindcast data from three leading operational centers as benchmarks. The data were obtained from the public database of the Subseasonal-to-Seasonal (S2S) Prediction Project, a joint initiative of the World Weather Research Programme (WWRP) and the World Climate Research Programme (WCRP)[60]. The specific models are as follows:

ECMWF: Ensemble hindcasts from the European Centre for Medium-Range Weather Forecasts were used as the primary dynamical benchmark, consisting of 11 members per initialization. The re-forecast frequency in the S2S archive varies over the study period (ranging from twice a week to daily or every two days).

CMA: Ensemble hindcasts from the China Meteorological Administration were also included, consisting of 4 members per initialization. Similar to the ECMWF dataset, the initialization frequency varies (e.g., daily or twice a week) depending on the archival period.

Since the CMA and ECMWF hindcasts are not initialized daily, their initialization dates for comparison with FM-Cast were chosen to be the closest available date to the target lead time.

## FM-Cast

Traditional diffusion models like DDPM[61], while capable of generating high-fidelity and diverse samples in fields such as image generation, suffer from low efficiency. Their sampling process follows a Markov chain that often requires thousands of iterative steps to produce a single sample. Although subsequent methods like DDIM[62], DPM-Solver[63], and DPM-Solver++[64] significantly accelerated sampling by reframing the generation process as an ordinary differential equation (ODE) solvable with dedicated numerical solvers, these solvers can be complex to implement, and the model training still relies on a predefined, step-by-step forward noising process. Flow Matching (FM) offers an elegant solution to these challenges as a more general generative framework[65-67]. FM directly learns the velocity field that connects the data and noise distributions. By designing this as a straight path, the model constructs a smoother, more direct transformation trajectory in high-dimensional space, enabling the generation of high-quality samples in just 10–20 ODE integration steps. Unlike diffusion models, FM does not require a prescribed multi-step noising process; instead, it directly constructs the learning target between data points and random noise, leading to a simpler and more stable training procedure.

Furthermore, FM is not restricted to Gaussian noise and can theoretically map between any two arbitrary distributions, making it a more general and direct extension of diffusion models[66].

The FM-Cast model (Fig. 1), based on Flow Matching and U-Net architecture[68], draws inspiration from the conditional generation architecture of Gencast[32], the U-Net implementation from the codebase of Zhang et al.[69], and the noise level injection scheme from Rombach et al.[70]. Similar to Gencast[32], FM-Cast learns the conditional probability distribution $P(X^{d+1}|X^{d}, X^{d-1})$, where the superscript $d$ denotes the time in days for the meteorological data, to distinguish it from the noise level $t$ in the flow matching process.

Fig. 1a illustrates the basic architecture of FM-Cast. The model takes three inputs: a linear interpolation $X_t = tX_1 + (1-t)X_0$ between random Gaussian noise $X_0$ and the target data $X_1$; the meteorological variables from the previous two days as the condition $\mathrm{Cond}$; and the noise level $t$. This noise level $t$ is injected into the normalization modules of each U-Net layer via an embedding, a design that allows the neural network to be fully aware of the current noise intensity. The U-Net architecture consists of a signal-noise encoder (four noise-aware convolutional downsampling blocks and a self-attention block), a condition encoder (four convolutional downsampling blocks), and a decoder (four noise-aware convolutional upsampling blocks). Skip connections merge the multi-level features from $X_t$ and the condition, feeding them into the corresponding decoder blocks to ensure a thorough fusion of noise and conditional information. The output of FM-Cast is the predicted velocity field between the data and noise distributions in the high-dimensional probability space. During training, the model is optimized by minimizing a loss function $\min \int_0^1 E[\|(X_1 - X_0) - v(X_t, t, \mathrm{cond})\|^2]dt$, ensuring that the predicted velocity field $v(X_t, t, \mathrm{Cond})$ closely approximates the true velocity field $X_1 - X_0$ across all noise levels $t$. After training, new samples are generated by solving the ordinary differential equation (ODE): $dX_t = v(t)dt,\ t \in [0,1]$, $\widehat{X_1} = \int_0^1 v(t)dt$.

Fig. 1b details the design of each U-Net module, which is adapted from the implementation in Zhang et al.[69]. We introduced two key modifications to better capture the physical processes relevant to SSWs.

First, we replaced all standard 2D convolution (Conv2d) layers with zonally periodic convolutions to correctly handle the periodic boundary conditions of atmospheric circulation in the longitudinal direction. Second, we replaced Batch Normalization (BatchNorm) layers with Group Normalization (GroupNorm) to enhance training stability and performance with small batch sizes. Furthermore, we conditioned the GroupNorm layers within the signal-noise encoder and the decoder to inject the noise level information from the flow matching process. The core of this mechanism is adaptive normalization: we first normalize the features using a standard GroupNorm layer but without its learnable affine transformation parameters (i.e., affine=False). The required scale (gamma) and shift (beta) parameters for the affine transformation are then dynamically predicted by a separate linear projection layer based on the current noise level embedding vector. Finally, the predicted gamma and beta are applied to the normalized feature map. Through this mechanism, the U-Net's encoder and decoder can dynamically adjust their feature responses according to the different stages of the generation process (i.e., different noise levels), achieving fine-grained control over the generation.

The FM-Cast model was trained with the following hyperparameters: the initial learning rate was set to 0.0001 and followed a step decay schedule, halving every 30 epochs. The batch size was 16 for the first 30 epochs and was subsequently reduced to 8. The model was trained for a total of 100 epochs. To enhance generalization, weight decay with a coefficient of 0.0001 was applied. The entire training process was conducted on a single NVIDIA A800 GPU and took approximately 4 hours for each period. Because generative models learn a velocity field in a high-dimensional probability space, the conventional validation loss does not fully reflect the physical realism or forecast skill of the generated samples. In image generation, metrics such as the Fréchet Inception Distance (FID) are widely used to evaluate sample quality[71], but these metrics rely on feature extractors pre-trained on natural images and are therefore not applicable to meteorological data. We therefore adopted a case-study-based approach to select the optimal model. Specifically, we used model weights from different checkpoints during training to generate forecasts for several key historical SSW events. By comparing the forecast performance on these critical events—assessing features such as the polar vortex morphology, intensity, and evolutionary

trend—we selected the model weights that yielded the best overall performance. In this study, the model from around the 60-70th epoch demonstrated the best comprehensive performance.

The inference process of FM-Cast is illustrated in Fig. 1c. Similar to Gencast[32], the model takes the meteorological variables from the previous time steps as a condition, along with a random Gaussian noise sample, to predict a velocity field. This field is then integrated with respect to the noise level $t$ from 0 to 1 to generate the forecast for the next time step, $X_1$. In this study, we use 20 integration steps. This new state is then fed back into the model as part of the condition in an autoregressive loop to generate longer-term forecasts. (For simplicity, the diagram shows only one day of input conditions and omits the detailed ODE integration process). Ensemble forecasts are produced by sampling from different initial Gaussian noise maps, a process that can be computed in parallel on a GPU. In the "perfect troposphere" experiment, after the initial conditions are set, the forecasted temperature and geopotential height at 850 hPa and 500 hPa are replaced with their true values at each step of the autoregressive loop. All other variables remain as forecasted. This procedure is designed to simulate the impact of perfect tropospheric forcing on the stratospheric forecast results.

## Evaluation metrics

**Assessment of Polar Vortex Morphology and Intensity**

To quantitatively evaluate the forecast skill of FM-Cast for the morphology and intensity of the polar vortex, we selected the polar vortex index, spatial Root Mean Squared Error (RMSE), and Anomaly Correlation Coefficient (ACC), following the metrics used in previous studies[16,44]. The polar vortex index is defined as the zonal mean of the zonal wind at 10 hPa and 60°N. The spatial RMSE and ACC are defined as follows:

$$\text{RMSE} = \sqrt{\frac{1}{60 \times 90} \sum_{\phi=1}^{60} \sum_{\lambda=1}^{90} \left(X_{\phi,\lambda}^{\text{Pred}} - X_{\phi,\lambda}^{\text{G.T.}}\right)^2}$$

$$\text{ACC} = \frac{\sum_{\phi=1}^{60} \sum_{\lambda=1}^{90} \left(X_{\phi,\lambda}^{\text{Pred}} - X_{\phi,\lambda}^{c}\right)\left(X_{\phi,\lambda}^{\text{G.T.}} - X_{\phi,\lambda}^{c}\right)}{\sqrt{\sum_{\phi=1}^{60} \sum_{\lambda=1}^{90} \left(X_{\phi,\lambda}^{\text{Pred}} - X_{\phi,\lambda}^{c}\right)^2 \sum_{\phi=1}^{60} \sum_{\lambda=1}^{90} \left(X_{\phi,\lambda}^{\text{G.T.}} - X_{\phi,\lambda}^{c}\right)^2}}$$

where $X_{\phi,\lambda}^{Pred}$ represents the forecasted variable field for a given day over the 30°N–89°N domain, $X_{\phi,\lambda}^{G.T.}$ is the corresponding ERA5 verification field, and $X_{\phi,\lambda}^{c}$ is the climatological mean field, which is calculated by averaging all data from November to March across all years.

**Definition of SSW Criteria and Ensemble Forecast Accuracy**

An SSW event is identified in the forecast when the zonal mean zonal wind at 10 hPa and 60°N reverses to easterlies (i.e., drops below 0 m/s)[72]. To account for the temporal uncertainties inherent in subseasonal forecasting, a prediction is classified as successful if the forecasted wind reversal occurs within a specific tolerance window relative to the observed onset date. We adopted a lead-time dependent tolerance threshold: for long-range forecasts (lead time $\geq$ 10 days), a prediction is considered successful if the onset falls within $\pm 5$ days of the actual date; for medium-range forecasts (lead time 6–9 days), the tolerance is tightened to $\pm 3$ days; and for short-range forecasts (lead time $\leq$ 5 days), a precise onset within $\pm 2$ days is required.

To quantify the SSW forecast probability, we employed two distinct methods:

Frequency Counting Method: The probability is calculated as the fraction of ensemble members that satisfy the SSW criterion on a given forecast day.

Gaussian PDF-based Method: To mitigate the issue of coarse probabilistic resolution caused by small ensemble sizes in the NWP hindcasts (e.g., step changes of 0.25 in probability for 4 members), we adopted a parametric approach. We assume that the zonal wind distribution across ensemble members at each lead time follows a normal (Gaussian) distribution. The Probability Density Function (PDF) is constructed using the instantaneous ensemble mean $\mu$ and standard deviation $\sigma$. The probability of an SSW $P_{SSW}$ is then calculated as the cumulative area of the PDF that falls below the threshold of 0 m/s:

$$P_{SSW} = \int_{-\infty}^{0} \frac{1}{\sigma\sqrt{2\pi}} e^{-\frac{(x-\mu)^2}{2\sigma^2}} dx$$

While the assumption of normality ignores potential multi-modal flow regimes, previous studies Karpechko et al. (2018)[46] have demonstrated that this method yields results comparable to other approaches for specific events, justifying its application for consistent comparison across models with varying ensemble sizes.

**Ensemble Subsampling for Fair Comparison**

While FM-Cast generates 50 ensemble members by default, the operational hindcasts available for comparison have limited sizes (e.g., 11 for ECMWF). To ensure a fair comparison in the direct time-series visualizations (e.g., Fig. 3), we selected the first 11 members from the 50 randomly generated ensemble members of FM-Cast to match the ECMWF ensemble size. However, for the calculation of probabilistic metrics (Accuracy, Reliability) and overall statistics (Fig. 2, Fig. 4), the full 50-member ensemble was utilized to leverage FM-Cast's full probabilistic capability, unless otherwise stated. The Gaussian PDF method described above ensures that probabilistic scores remain comparable despite differences in ensemble size.

## Reliability Diagram Construction

To assess the calibration of the probabilistic forecasts, we constructed a reliability diagram following the approach described in Karpechko et al. (2018)[46]. We aggregated all daily forecasts from the three testing periods (1998–2024). The forecasted SSW probabilities (derived from the Gaussian PDF method) were grouped into 10 equally spaced bins (width of 0.1) ranging from 0 to 1. For each bin, we calculated the forecast probability (the average probability of forecasts falling within that bin) and the observed frequency (the fraction of those forecast days on which an SSW event actually occurred).

In this diagram, a curve lying close to the 1:1 diagonal indicates a reliable (well-calibrated) forecast system. Points below the diagonal suggest overconfidence (forecasting a higher probability than the actual occurrence rate), while points above indicate underconfidence. This metric allows us to evaluate the model's performance generally, ensuring that it correctly assigns low probabilities during non-SSW periods (minimizing false alarms) and high probabilities only when justified.

## Data Availability

The ERA5 reanalysis data are accessible through the Copernicus Climate Data Store at https://doi.org/10.24381/cds.bd0915c6 or https://cds.climate.copernicus.eu/. The Subseasonal-to-Seasonal (S2S) Prediction Project can be retrieved from https://apps.ecmwf.int/datasets/.

## Code Availability

The FM-Cast source code will be deposited in a public GitHub repository and made available at the time of publication.

## Figures

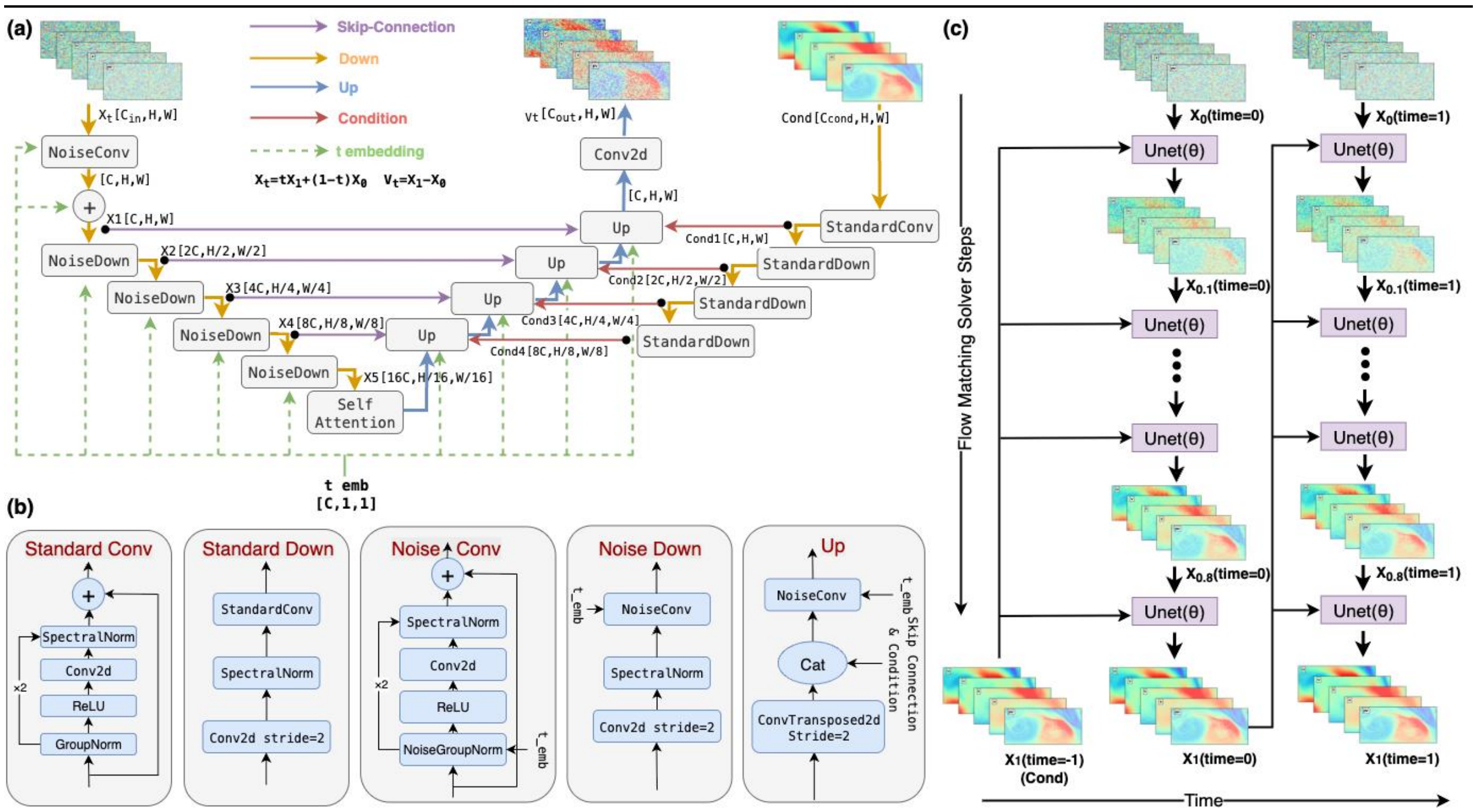

**Fig. 1 | The architecture of the FM-Cast model. a** Architectural details of FM-Cast. The model employs a U-Net architecture, which primarily consists of a signal-noise encoder (four noise-aware convolutional downsampling blocks and a self-attention block), a condition encoder (four convolutional downsampling blocks), and a decoder (four noise-aware convolutional upsampling blocks). During training, FM-Cast takes a linear interpolation $X_t$ between random Gaussian noise $X_0$ and the target data $X_1$, the meteorological variables from the previous two days as the condition $\mathrm{Cond}$, and the noise level $t$ as inputs. It outputs a prediction of the velocity field in the high-dimensional probability space. **b** Design details of the modules within FM-Cast. **c** The autoregressive inference process of FM-Cast. Sampling from the conditional probability is equivalent to computing a numerical integration, where the subscript $t$ denotes the noise level. The velocity field $v_t$ is obtained from the trained FM-Cast network with $X_t$ and $\mathrm{Cond}$ as inputs. In the diagram, the subscript of $X$ represents the noise level, while (time) refers to the time in days for the meteorological variable. For simplicity, the input condition is shown for only one day, and the ODE integration process is not explicitly drawn. After a forecast sample is obtained by numerically integrating from the initial Gaussian noise, it is used as a new condition, and the sampling

operation is repeated autoregressively to generate longer-term forecasts. (The schematic for the input variables shows only 5 dimensions—representing zonal wind, meridional wind, temperature, geopotential height, and potential vorticity—whereas the actual model uses 33 dimensions. For full details of FM-Cast, see Methods).

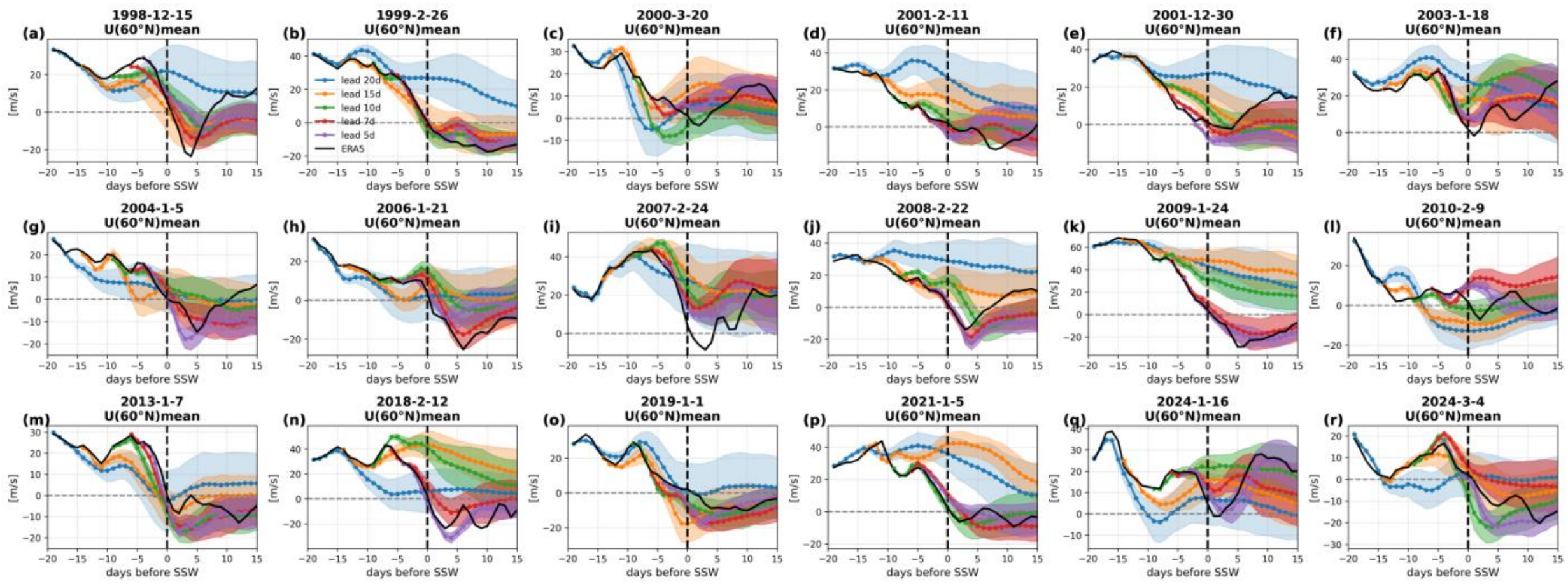

**Fig. 2 | FM-Cast forecasts of the polar vortex index (10 hPa, 60 ˚ N zonal-mean zonal wind) for the 18 major SSW events from 1998 – 2024 with the ERA5 verification (black solid line) .** The x-axis represents the number of days relative to the SSW onset date (negative values are days before onset, positive values are days after, and the vertical dashed line marks the onset day). The colored dotted lines represent the 50-member ensemble mean, while the shaded areas indicate the ensemble standard deviation. Different colors correspond to different forecast lead times (blue: 20 days, orange: 15 days, green: 10 days, red: 7 days, purple: 5 days). The horizontal gray dashed line indicates the zero-wind line.

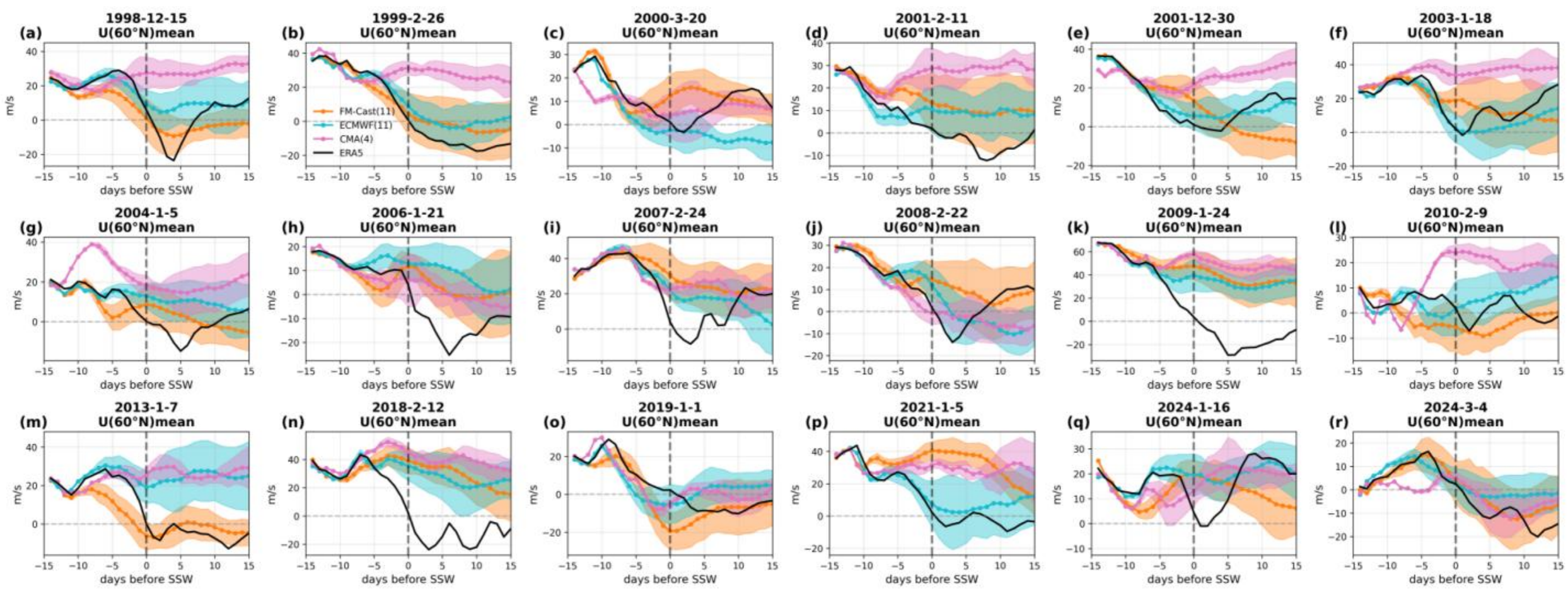

**Fig. 3 | Forecast skill comparison between FM-Cast and leading NWP models for the polar vortex index (10 hPa, 60 ° N zonal mean zonal wind) at a lead time of 15 days for 18 SSW events.** The x-axis represents the number of days relative to the SSW onset date (negative values are days before onset, positive values are days after, and the vertical dashed line marks the onset day). The black solid line represents the ERA5 verification. The orange dotted line represents the FM-Cast ensemble mean (11 members), the cyan dotted line represents the ECMWF ensemble mean (11 members), and the pink dotted line represents the CMA ensemble mean (4 members). Shaded areas indicate the ensemble standard deviation. The horizontal gray dashed line indicates the zero-wind line.

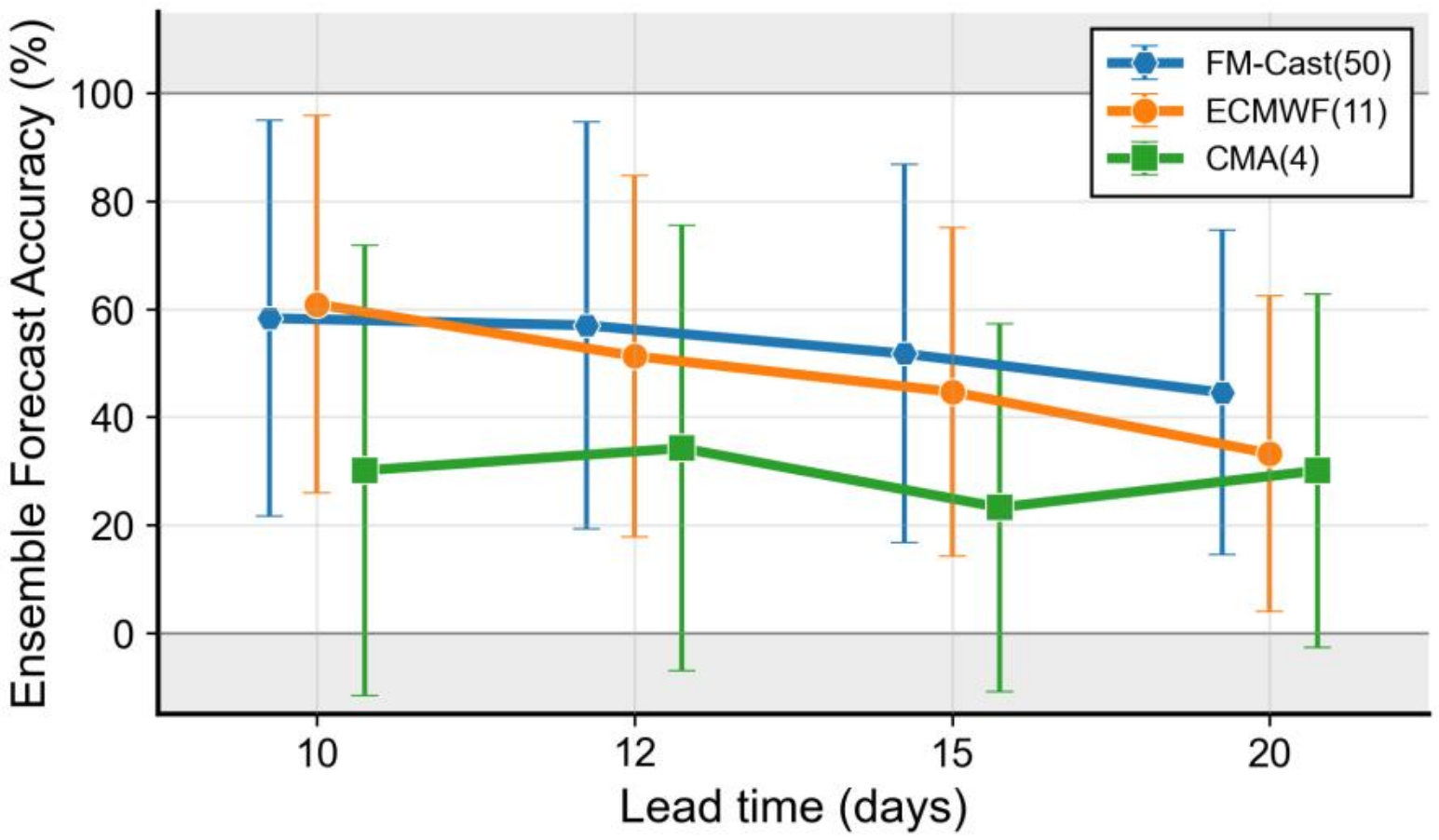

**Fig. 4 | Comparison of average ensemble forecast accuracy for the 18 major SSW events from 1998–2024 between FM-Cast and the leading NWP models.** The y-axis represents the ensemble

forecast accuracy calculated using the Gaussian PDF-based Method (50 ensemble members for FM-Cast versus 11 for ECMWF and 4 ensemble members for CMA), and the x-axis represents the forecast lead time (days). Error bars indicate the standard deviation across the 18 events. Ensemble forecast accuracy are robust to both the choice of probability calculation method and the ensemble size (see Fig. S7).

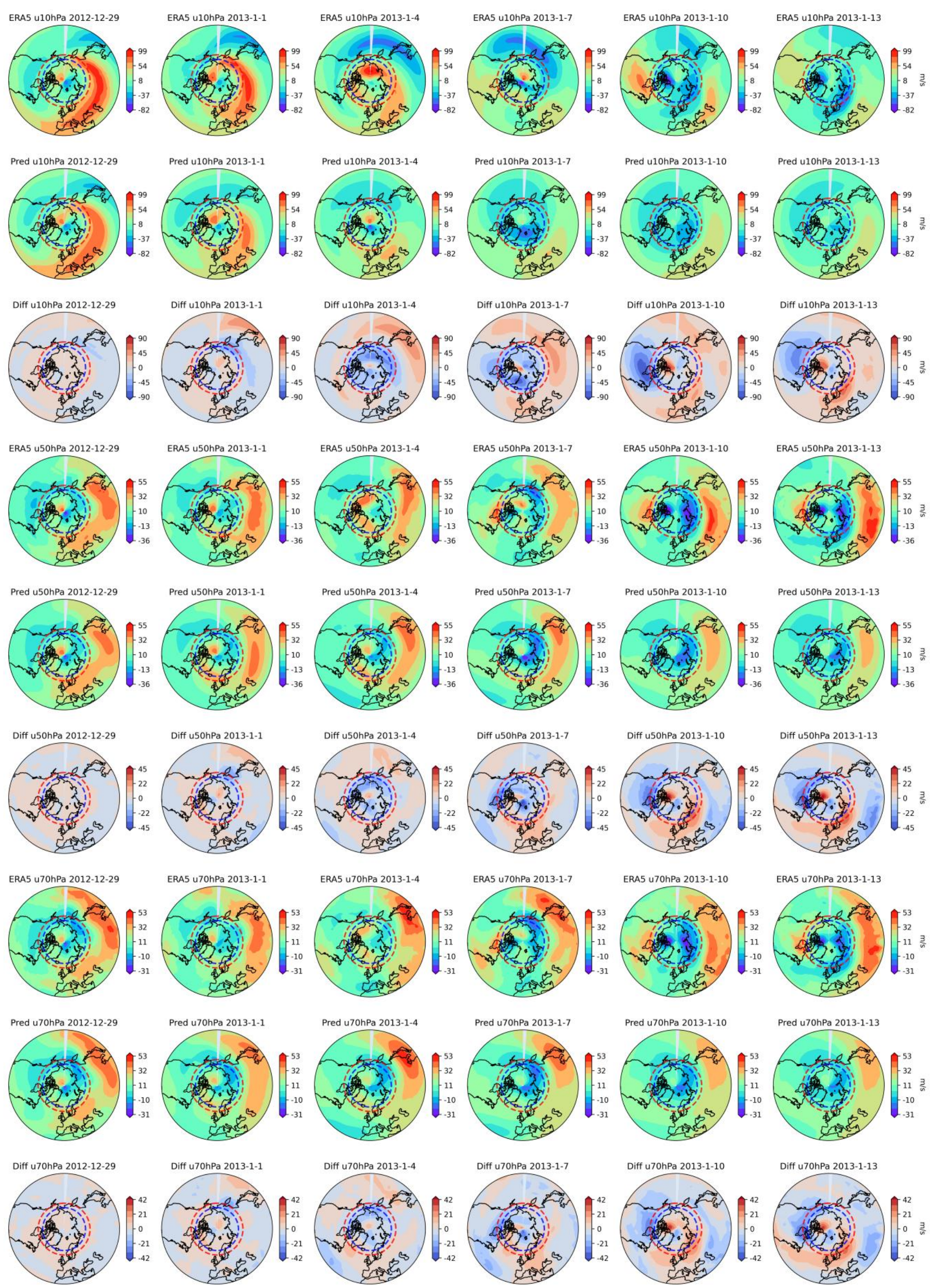

**Fig. 5 | Forecast performance of FM-Cast for the zonal wind field during a split-type SSW event (7 January 2013).** Shown are forecasts from 70 hPa to 10 hPa over the 30 ˚ N – 89 ˚ N domain. The forecast was initialized on 25 and 26 December 2012. The columns from left to right are separated by 3-

day intervals. The first, fourth, and seventh rows display the ERA5 field, while the second, fifth, and eighth rows display the corresponding forecasts (Pred.), the third, sixth, ninth rows display the difference between pred and ERA5, for the 10 hPa, 50 hPa, and 70 hPa levels, respectively. The results shown are the ensemble mean of the 50-member ensemble.

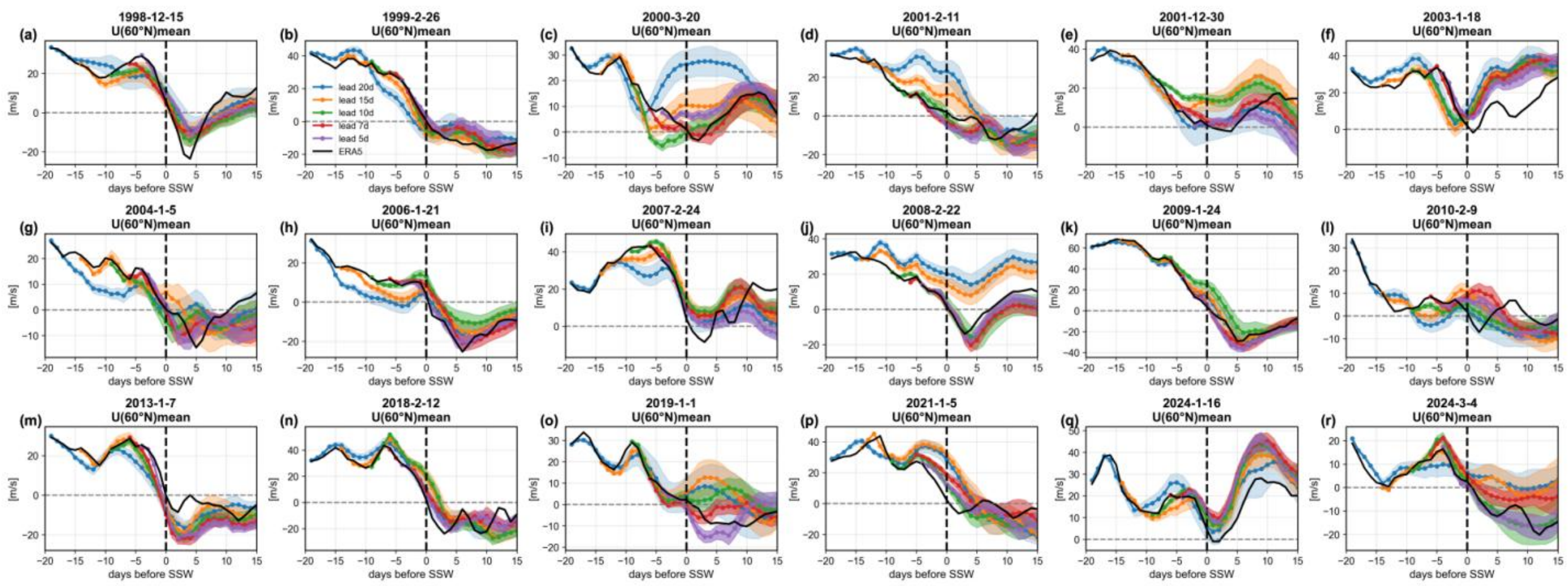

**Fig. 6 | FM-Cast forecasts of the polar vortex index (10 hPa, 60 ° N zonal-mean zonal wind) for the 18 major SSW events from 1998 – 2024 with the ERA5 verification (black solid line) under "perfect troposphere" forcing.** The x-axis represents the number of days relative to the SSW onset date (negative values are days before onset, positive values are days after, and the vertical dashed line marks the onset day). The colored dotted lines represent the 50-member ensemble mean, while the shaded areas indicate the ensemble standard deviation. Different colors correspond to different forecast lead times. The horizontal gray dashed line indicates the zero-wind line.

## References

1. Baldwin, M. P., Birner, T., Brasseur, G., Burrows, J., Butchart, N., Garcia, R., ... & Scaife, A. A. (2019). 100 years of progress in understanding the stratosphere and mesosphere. Meteorological Monographs, 59, 27-1.
2. Bazilevskaya, G. A., Usoskin, I. G., Flückiger, E. O., Harrison, R. G., Desorgher, L., Bütikofer, R., ... & Kovaltsov, G. A. (2008). Cosmic ray induced ion production in the atmosphere. Space Science Reviews, 137(1), 149-173.
3. Haynes, P. H., McIntyre, M. E., Shepherd, T. G., Marks, C. J., & Shine, K. P. (1991). On the "downward control" of extratropical diabatic circulations by eddy-induced mean zonal forces. Journal of Atmospheric Sciences, 48(4), 651-678.
4. Thompson, D. W., & Wallace, J. M. (1998). The Arctic Oscillation signature in the wintertime geopotential height and temperature fields. Geophysical research letters, 25(9), 1297-1300.
5. Baldwin, M. P., & Dunkerton, T. J. (1999). Propagation of the Arctic Oscillation from the stratosphere to the troposphere. Journal of Geophysical Research: Atmospheres, 104(D24), 30937-30946.
6. Kidston, J., Scaife, A. A., Hardiman, S. C., Mitchell, D. M., Butchart, N., Baldwin, M. P., & Gray, L. J. (2015). Stratospheric influence on tropospheric jet streams, storm tracks and surface weather. Nature Geoscience, 8(6), 433-440.
7. Baldwin, M. P., & Dunkerton, T. J. (2001). Stratospheric harbingers of anomalous weather regimes. Science, 294(5542), 581-584.
8. Baldwin, M. P., Ayarzagüena, B., Birner, T., Butchart, N., Butler, A. H., Charlton-Perez, A. J., ... & Pedatella, N. M. (2021). Sudden stratospheric warmings. Reviews of Geophysics, 59(1), e2020RG000708.

9. Charlton-Perez, A. J., Ferranti, L., & Lee, R. W. (2018). The influence of the stratospheric state on North Atlantic weather regimes. Quarterly Journal of the Royal Meteorological Society, 144(713), 1140-1151.

10. Domeisen, D. I., Grams, C. M., & Papritz, L. (2020). The role of North Atlantic–European weather regimes in the surface impact of sudden stratospheric warming events. Weather and Climate Dynamics, 1(2), 373-388.

11. King, A. D., Butler, A. H., Jucker, M., Earl, N. O., & Rudeva, I. (2019). Observed relationships between sudden stratospheric warmings and European climate extremes. Journal of Geophysical Research: Atmospheres, 124(24), 13943-13961.

12. Harnik, N., Messori, G., Caballero, R. & Feldstein, S. B. The Circumglobal North American wave pattern and its relation to cold events in eastern North America. Geophysical Research Letters 43, 11,015-011,023, doi:https://doi.org/10.1002/2016GL070760 (2016).

13. Shaw, T. A. & Perlwitz, J. The Life Cycle of Northern Hemisphere Downward Wave Coupling between the Stratosphere and Troposphere. Journal of Climate 26, 1745-1763, doi:https://doi.org/10.1175/JCLI-D-12-00251.1 (2013).

14. 28 Messori, G., Kretschmer, M., Lee, S. H. & Wendt, V. Stratospheric downward wave reflection events modulate North American weather regimes and cold spells. Weather Clim. Dynam. 3, 1215-1236, doi:10.5194/wcd-3-1215-2022 (2022).

15. Domeisen, D. I., Butler, A. H., Charlton-Perez, A. J., Ayarzagüena, B., Baldwin, M. P., Dunn-Sigouin, E., ... & Taguchi, M. (2020). The role of the stratosphere in subseasonal to seasonal prediction: 1. Predictability of the stratosphere. Journal of Geophysical Research: Atmospheres, 125(2), e2019JD030920.

16. Rao, J., Garfinkel, C. I., Chen, H., & White, I. P. (2019). The 2019 new year stratospheric sudden warming and its real-time predictions in multiple S2S models. Journal of Geophysical Research: Atmospheres, 124(21), 11155-11174.

17. Noguchi, S., Mukougawa, H., Kuroda, Y., Mizuta, R., Yabu, S., & Yoshimura, H. (2016). Predictability of the stratospheric polar vortex breakdown: An ensemble reforecast experiment for the splitting event in January 2009. Journal of Geophysical Research: Atmospheres, 121(7), 3388-3404.

18. Rao, J., Ren, R., Chen, H., Yu, Y., & Zhou, Y. (2018). The stratospheric sudden warming event in February 2018 and its prediction by a climate system model. Journal of Geophysical Research: Atmospheres, 123(23), 13-332.

19. Karpechko, A. Y., Charlton-Perez, A., Balmaseda, M., Tyrrell, N., & Vitart, F. (2018). Predicting sudden stratospheric warming 2018 and its climate impacts with a multimodel ensemble. Geophysical Research Letters, 45(24), 13-538.

20. Lee, S. H., Charlton-Perez, A. J., Furtado, J. C., & Woolnough, S. J. (2019). Abrupt stratospheric vortex weakening associated with North Atlantic anticyclonic wave breaking. Journal of Geophysical Research: Atmospheres, 124(15), 8563-8575.

21. Wu, R. W. Y., Chiodo, G., Polichtchouk, I., & Domeisen, D. I. (2024). Tropospheric links to uncertainty in stratospheric subseasonal predictions. Atmospheric Chemistry and Physics, 24(21), 12259-12275.

22. Lorenz, E. N. (2017). Deterministic nonperiodic flow 1. In Universality in Chaos, 2nd edition (pp. 367-378). Routledge.

23. Eichinger, R., Garny, H., Šácha, P., Danker, J., Dietmüller, S., & Oberländer-Hayn, S. (2020). Effects of missing gravity waves on stratospheric dynamics; part 1: climatology. Climate Dynamics, 54(5), 3165-3183.

24. Palmer, T. N., Buizza, R., Doblas-Reyes, F., Jung, T., Leutbecher, M., Shutts, G. J., ... & Weisheimer, A. (2009). Stochastic parametrization and model uncertainty.

25. Bauer, P., Dueben, P. D., Hoefler, T., Quintino, T., Schulthess, T. C., & Wedi, N. P. (2021). The digital revolution of Earth-system science. Nature Computational Science, 1(2), 104-113.

26. Schulthess, T. C., Bauer, P., Wedi, N., Fuhrer, O., Hoefler, T., & Schär, C. (2018). Reflecting on the goal and baseline for exascale computing: a roadmap based on weather and climate simulations. Computing in Science & Engineering, 21(1), 30-41.
27. Leutbecher, M., Lock, S. J., Ollinaho, P., Lang, S. T., Balsamo, G., Bechtold, P., ... & Weisheimer, A. (2017). Stochastic representations of model uncertainties at ECMWF: State of the art and future vision. Quarterly Journal of the Royal Meteorological Society, 143(707), 2315-2339.
28. Ham, Y. G., Kim, J. H., & Luo, J. J. (2019). Deep learning for multi-year ENSO forecasts. Nature, 573(7775), 568-572.
29. Bi, K., Xie, L., Zhang, H., Chen, X., Gu, X., & Tian, Q. (2023). Accurate medium-range global weather forecasting with 3D neural networks. Nature, 619(7970), 533-538.
30. Li, L., Carver, R., Lopez-Gomez, I., Sha, F., & Anderson, J. (2024). Generative emulation of weather forecast ensembles with diffusion models. Science Advances, 10(13), eadk4489.
31. Lam, R., Sanchez-Gonzalez, A., Willson, M., Wirnsberger, P., Fortunato, M., Alet, F., ... & Battaglia, P. (2023). Learning skillful medium-range global weather forecasting. Science, 382(6677), 1416-1421.
32. Price, I., Sanchez-Gonzalez, A., Alet, F., Andersson, T. R., El-Kadi, A., Masters, D., ... & Willson, M. (2025). Probabilistic weather forecasting with machine learning. Nature, 637(8044), 84-90.
33. Ravuri, S., Lenc, K., Willson, M., Kangin, D., Lam, R., Mirowski, P., ... & Mohamed, S. (2021). Skilful precipitation nowcasting using deep generative models of radar. Nature, 597(7878), 672-677.
34. Das, P., Posch, A., Barber, N., Hicks, M., Duffy, K., Vandal, T., ... & Ganguly, A. R. (2024). Hybrid physics-AI outperforms numerical weather prediction for extreme precipitation nowcasting. npj Climate and Atmospheric Science, 7(1), 282.
35. Chen, L., Du, F., Hu, Y., Wang, Z., & Wang, F. (2023, June). Swinrdm: integrate swinrnn with diffusion model towards high-resolution and high-quality weather forecasting. In Proceedings of the AAAI Conference on Artificial Intelligence (Vol. 37, No. 1, pp. 322-330).

36. Yoon, D., Seo, M., Kim, D., Choi, Y., & Cho, D. (2024, September). Probabilistic weather forecasting with deterministic guidance-based diffusion model. In European Conference on Computer Vision (pp. 108-124). Cham: Springer Nature Switzerland.

37. Zhong, X., Chen, L., Liu, J., Lin, C., Qi, Y., & Li, H. (2024). FuXi-Extreme: Improving extreme rainfall and wind forecasts with diffusion model. Science China Earth Sciences, 67(12), 3696-3708.

38. Rühling Cachay, S., Henn, B., Watt-Meyer, O., Bretherton, C. S., & Yu, R. (2024). Probablistic Emulation of a Global Climate Model with Spherical DYffusion. Advances in Neural Information Processing Systems, 37, 127610-127644.

39. Nai, C., Pan, B., Chen, X., Tang, Q., Ni, G., Duan, Q., ... & Liu, X. (2024). Reliable precipitation nowcasting using probabilistic diffusion models. Environmental Research Letters, 19(3), 034039.

40. Yang, S., Nai, C., Liu, X., Li, W., Chao, J., Wang, J., ... & Pan, B. (2025). Generative assimilation and prediction for weather and climate. arXiv preprint arXiv:2503.03038.

41. Peng, K., Cao, X., Liu, B., Guo, Y., Xiao, C., & Tian, W. (2021). Polar vortex multi-day intensity prediction relying on new deep learning model: A combined convolution neural network with long short-term memory based on Gaussian smoothing method. Entropy, 23(10), 1314.

42. Wu, Z., Beucler, T., Székely, E., Ball, W. T., & Domeisen, D. I. (2022). Modeling stratospheric polar vortex variation and identifying vortex extremes using explainable machine learning. Environmental Data Science, 1, e17.

43. Chen, Y. C., Liang, Y. C., Wu, C. M., Huang, J. D., Lee, S. H., Wang, Y., & Zeng, Y. J. (2024). Exploiting a variational auto-encoder to represent the evolution of sudden stratospheric warmings. Environmental Research: Climate, 3(2), 025006.

44. Du, Y., Zhang, J., Cheng, X., Lu, Y., Li, D., & Tian, W. (2025). Predicting sudden stratospheric warmings using video prediction methods. Geophysical Research Letters, 52(8), e2024GL113993.

45. Taguchi, M. (2020). Verification of subseasonal-to-seasonal forecasts for major stratospheric sudden warmings in northern winter from 1998/99 to 2012/13. Advances in Atmospheric Sciences, 37(3), 250-258.

46. Karpechko, A. Y. (2018). Predictability of sudden stratospheric warmings in the ECMWF extended-range forecast system. Monthly Weather Review, 146(4), 1063-1075.

47. Rao, J., Garfinkel, C. I., & White, I. P. (2020). Predicting the downward and surface influence of the February 2018 and January 2019 sudden stratospheric warming events in subseasonal to seasonal (S2S) models. Journal of Geophysical Research: Atmospheres, 125(2), e2019JD031919.

48. Rao, J., Garfinkel, C. I., Wu, T., Lu, Y., Lu, Q., & Liang, Z. (2021). The January 2021 Sudden Stratospheric Warming and Its Prediction in Subseasonal to Seasonal Models. Journal of Geophysical Research: Atmospheres, 126(21), e2021JD035057. https://doi.org/10.1029/2021JD035057

49. Lee, S. H., Butler, A. H., & Manney, G. L. (2025). Two major sudden stratospheric warmings during winter 2023/2024. Weather, 80(2), 45–53. https://doi.org/10.1002/wea.7656

50. Chwat, D., Garfinkel, C. I., Chen, W., & Rao, J. (2022). Which Sudden Stratospheric Warming Events Are Most Predictable? Journal of Geophysical Research: Atmospheres, 127(18), e2022JD037521. https://doi.org/10.1029/2022JD037521

51. Bauer, P., Thorpe, A., & Brunet, G. (2015). The quiet revolution of numerical weather prediction. Nature, 525(7567), 47–55. https://doi.org/10.1038/nature14956

52. Lorenz, E. N. (1969). The predictability of a flow which possesses many scales of motion. Tellus, 21(3), 289-307.

53. Tripathi, O. P., Baldwin, M., Charlton-Perez, A., Charron, M., Cheung, J. C., Eckermann, S. D., ... & Stockdale, T. (2016). Examining the predictability of the stratospheric sudden warming of January 2013 using multiple NWP systems. Monthly Weather Review, 144(5), 1935-1960.

54. Yang, P., Bao, M., Ren, X., & Tan, X. (2023). The Role of Vortex Preconditioning in Influencing the Occurrence of Strong and Weak Sudden Stratospheric Warmings in ERA5 Reanalysis. https://doi.org/10.1175/JCLI-D-22-0319.1

55. Kent, C., Scaife, A. A., Seviour, W. J. M., Dunstone, N., Smith, D., & Smout-Day, K. (2023). Identifying Perturbations That Tipped the Stratosphere Into a Sudden Warming During January 2013. Geophysical Research Letters, 50(24), e2023GL106288. https://doi.org/10.1029/2023GL106288

56. Lawrence, Z. D., Abalos, M., Ayarzagüena, B., Barriopedro, D., Butler, A. H., Calvo, N., ... & Wu, R. W. Y. (2022). Quantifying stratospheric biases and identifying their potential sources in subseasonal forecast systems. Weather and Climate Dynamics Discussions, 2022, 1-37.

57. Garfinkel, C. I., Lawrence, Z. D., Butler, A. H., Dunn-Sigouin, E., Erner, I., Karpechko, A. Y., ... & Wu, R. W. Y. (2025). A process-based evaluation of biases in extratropical stratosphere–troposphere coupling in subseasonal forecast systems. Weather and Climate Dynamics, 6(1), 171-195.

58. Simmons, A., Soci, C., Nicolas, J., Bell, B., Berrisford, P., Dragani, R., ... & Schepers, D. (2020). Global stratospheric temperature bias and other stratospheric aspects of ERA5 and ERA5. 1.

59. Hersbach, H., Bell, B., Berrisford, P., Biavati, G., Horányi, A., Muñoz Sabater, J., Nicolas, J., Peubey, C., Radu, R., Rozum, I., Schepers, D., Simmons, A., Soci, C., Dee, D., Thépaut, J-N. (2023): ERA5 monthly averaged data on pressure levels from 1940 to present. Copernicus Climate Change Service (C3S) Climate Data Store (CDS), DOI: 10.24381/cds.6860a573 (Accessed on 04-10-2025)

60. Vitart, F., Ardilouze, C., Bonet, A., Brookshaw, A., Chen, M., Codorean, C., ... & Zhang, L. (2017). The subseasonal to seasonal (S2S) prediction project database. Bulletin of the American Meteorological Society, 98(1), 163-173.

61. Ho, J., Jain, A., & Abbeel, P. (2020). Denoising diffusion probabilistic models. Advances in neural information processing systems, 33, 6840-6851.

62. Song, J., Meng, C., & Ermon, S. (2020). Denoising diffusion implicit models. arXiv preprint arXiv:2010.02502.

63. Lu, C., Zhou, Y., Bao, F., Chen, J., Li, C., & Zhu, J. (2022). Dpm-solver: A fast ode solver for diffusion probabilistic model sampling in around 10 steps. Advances in neural information processing systems, 35, 5775-5787.

64. Lu, C., Zhou, Y., Bao, F., Chen, J., Li, C., & Zhu, J. (2025). Dpm-solver++: Fast solver for guided sampling of diffusion probabilistic models. Machine Intelligence Research, 1-22.

65. Liu, X., Gong, C., & Liu, Q. (2022). Flow straight and fast: Learning to generate and transfer data with rectified flow. arXiv preprint arXiv:2209.03003.

66. Lipman, Y., Chen, R. T., Ben-Hamu, H., Nickel, M., & Le, M. (2022). Flow matching for generative modeling. arXiv preprint arXiv:2210.02747.

67. Albergo, M. S., & Vanden-Eijnden, E. (2022). Building normalizing flows with stochastic interpolants. arXiv preprint arXiv:2209.15571.

68. Ronneberger, O., Fischer, P., & Brox, T. (2015, October). U-net: Convolutional networks for biomedical image segmentation. In International Conference on Medical image computing and computer-assisted intervention (pp. 234-241). Cham: Springer international publishing.

69. Zhang, Y., Long, M., Chen, K., Xing, L., Jin, R., Jordan, M. I., & Wang, J. (2023). Skilful nowcasting of extreme precipitation with NowcastNet. Nature, 619(7970), 526-532.

70. Rombach, R., Blattmann, A., Lorenz, D., Esser, P., & Ommer, B. (2022). High-resolution image synthesis with latent diffusion models. In Proceedings of the IEEE/CVF conference on computer vision and pattern recognition (pp. 10684-10695).

71. Heusel, M., Ramsauer, H., Unterthiner, T., Nessler, B., & Hochreiter, S. (2017). Gans trained by a two time-scale update rule converge to a local nash equilibrium. Advances in neural information processing systems, 30.

72. Charlton, A. J., & Polvani, L. M. (2007). A New Look at Stratospheric Sudden Warmings. Part I: Climatology and Modeling Benchmarks. Journal of Climate, 20(3), 449-469.

## Acknowledgements

We would like to thank Professor Yan Xia for his insightful discussions.

## Funding

This work was supported by the National Key R&D Program of China (Grant No. 2023YFE0109000); the National Natural Science Foundation of China Grant No. 12135003 and 42375070; and the National Key R&D Program of China (Grant No. 2024YFF0809004).

## Ethics declarations

---

### Competing interests

The authors declare no conflicts of interest.

# Supplementary Information

## A. Supplementary Figures

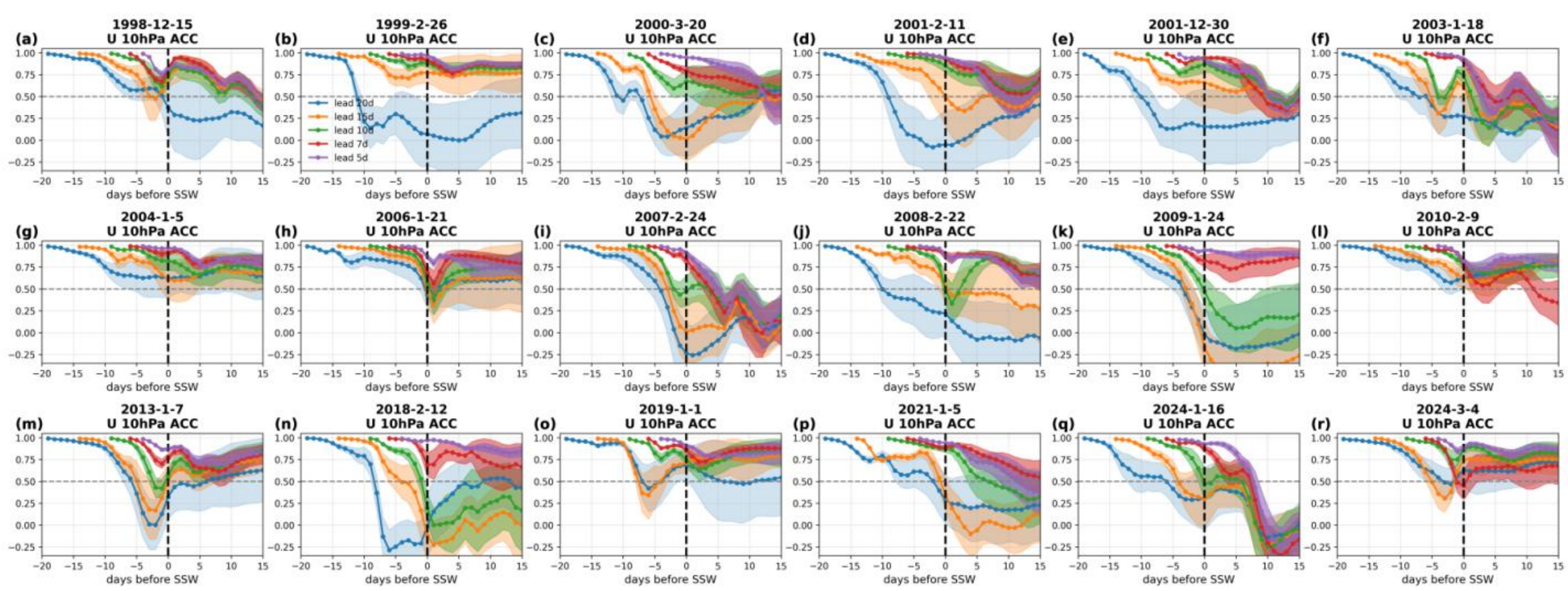

**Fig. S1 | Evolution of the Anomaly Correlation Coefficient (ACC) for 10 hPa zonal wind (U10) during the 18 major SSW events from 1998–2024.** The x-axis represents the number of days relative to the SSW onset date (negative values denote days before onset, positive values denote days after, and the vertical black dashed line marks the onset day). The colored solid lines with markers represent the 50-member ensemble mean ACC for different forecast lead times (blue: 20 days, orange: 15 days, green: 10 days, red: 7 days, purple: 5 days), while the shaded areas indicate the ensemble standard deviation. The horizontal gray dashed line indicates the ACC threshold of 0.5, often used as a limit for useful prediction skill.

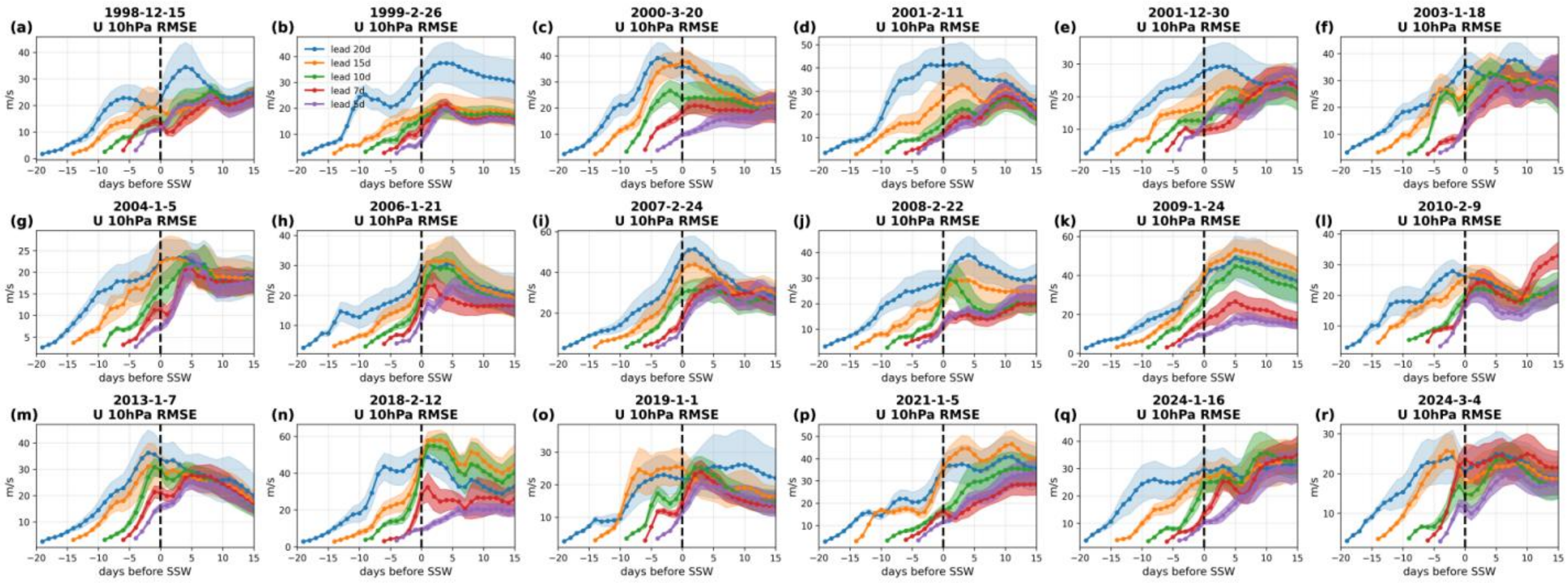

**Fig. S2 | Evolution of the Root Mean Squared Error (RMSE) for 10 hPa zonal wind (U10, 30°N–89°N) during the 18 major SSW events from 1998–2024.** The x-axis represents the number of days relative to the SSW onset date (negative values denote days before onset, positive values denote days after, and the vertical black dashed line marks the onset day). The colored solid lines with markers represent the 50-member ensemble mean RMSE for different forecast lead times (blue: 20 days, orange: 15 days, green: 10 days, red: 7 days, purple: 5 days), while the shaded areas indicate the ensemble standard deviation.

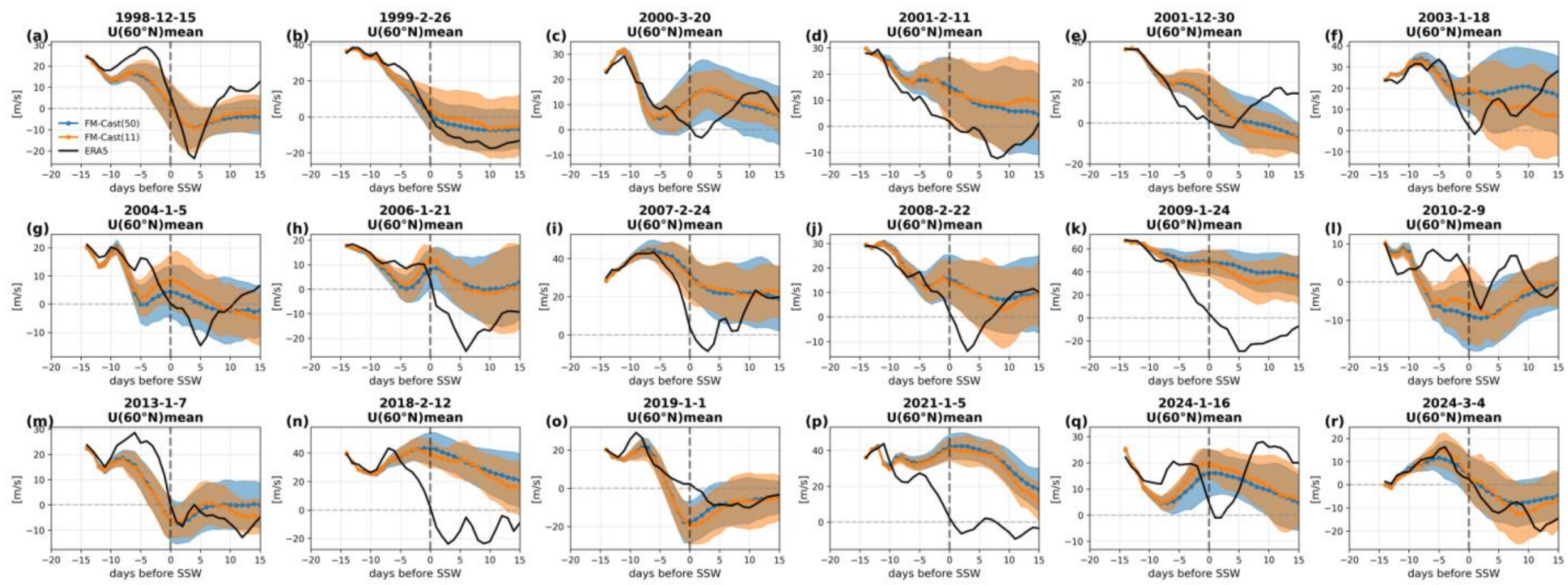

**Fig. S3 | Sensitivity analysis of FM-Cast ensemble size for the polar vortex index forecast.** Comparison of FM-Cast forecasts of the polar vortex index (10 hPa, 60°N zonal-mean zonal wind) at a lead time of 15 days for the 18 major SSW events from 1998–2024 using different ensemble sizes. The black solid line represents the ERA5 verification. The blue dotted line represents the ensemble mean of the full 50-member ensemble, with the blue shaded area indicating its standard deviation. The orange dotted line represents the ensemble mean of a subset consisting of the first 11 members selected from the 50-member ensemble, with the orange shaded area indicating its standard deviation. The x-axis represents the number of days relative to the SSW onset date (negative values are days before onset, positive values are days after, and the vertical dashed line marks the onset day). The horizontal gray dashed line indicates the zero-wind line. The overlapping trajectories and spreads demonstrate that the ensemble mean performance of FM-Cast is relatively insensitive to the reduction in ensemble size.

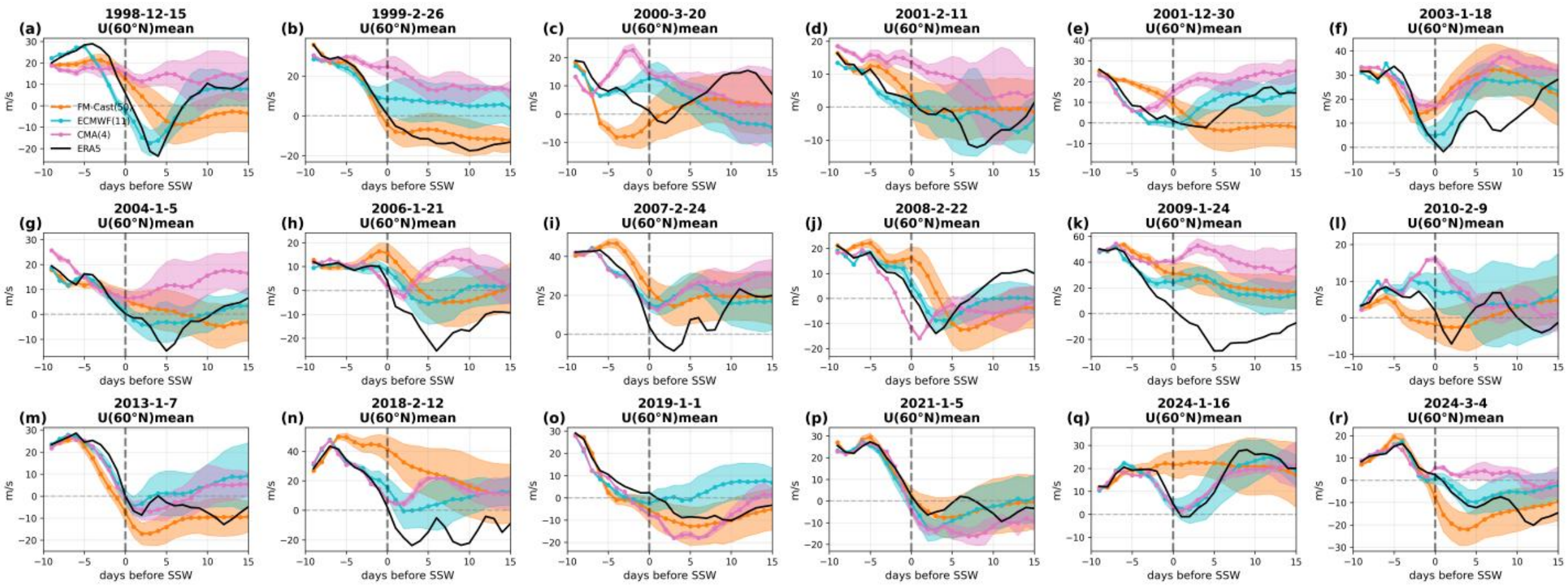

**Fig. S4 | Forecast skill comparison between FM-Cast and leading NWP models for the polar vortex index (10 hPa, 60°N zonal mean zonal wind) at a lead time of 10 days for 18 SSW events.** The x-axis represents the number of days relative to the SSW onset date (negative values are days before onset, positive values are days after, and the vertical dashed line marks the onset day). The black solid line represents the ERA5 verification. The orange dotted line represents the FM-Cast ensemble mean (11

members), the cyan dotted line represents the ECMWF ensemble mean (11 members), and the pink dotted line represents the CMA ensemble mean (4 members). Shaded areas indicate the ensemble standard deviation. The horizontal gray dashed line indicates the zero-wind line.

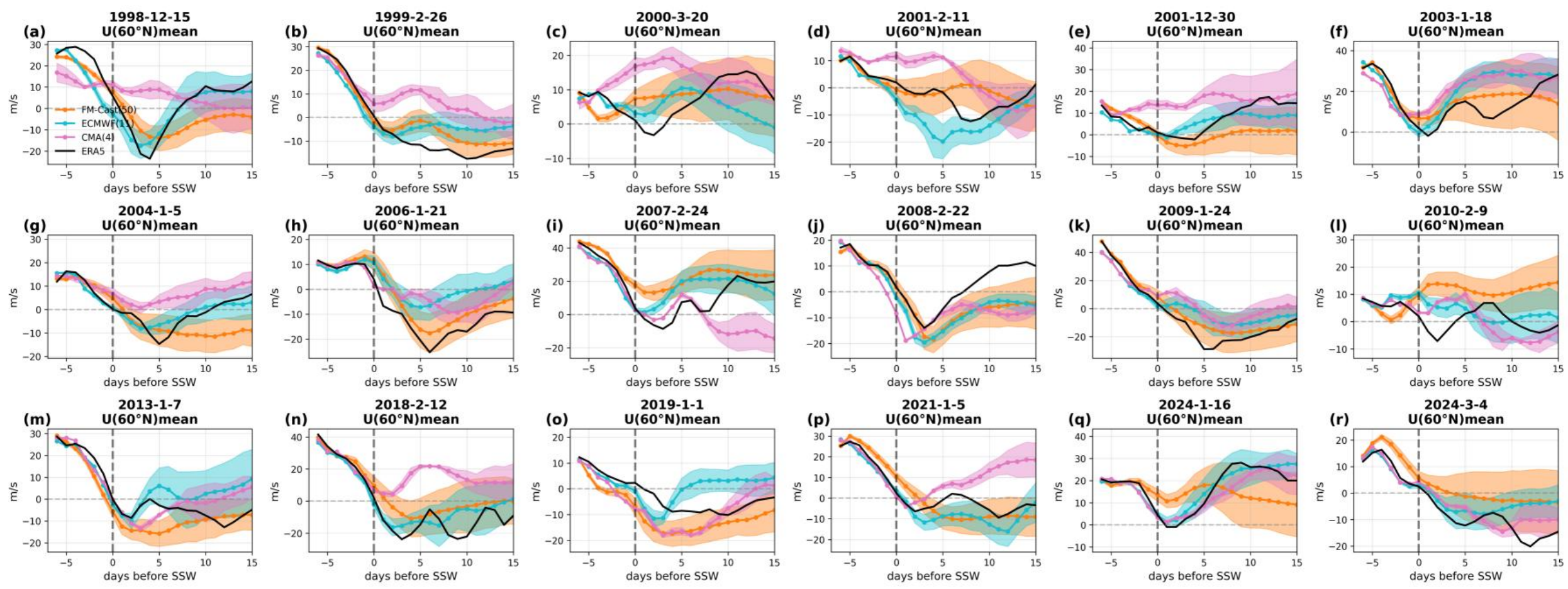

**Fig. S5 | Forecast skill comparison between FM-Cast and leading NWP models for the polar vortex index (10 hPa, 60°N zonal mean zonal wind) at a lead time of 7 days for 18 SSW events.** Same as Fig. S4.

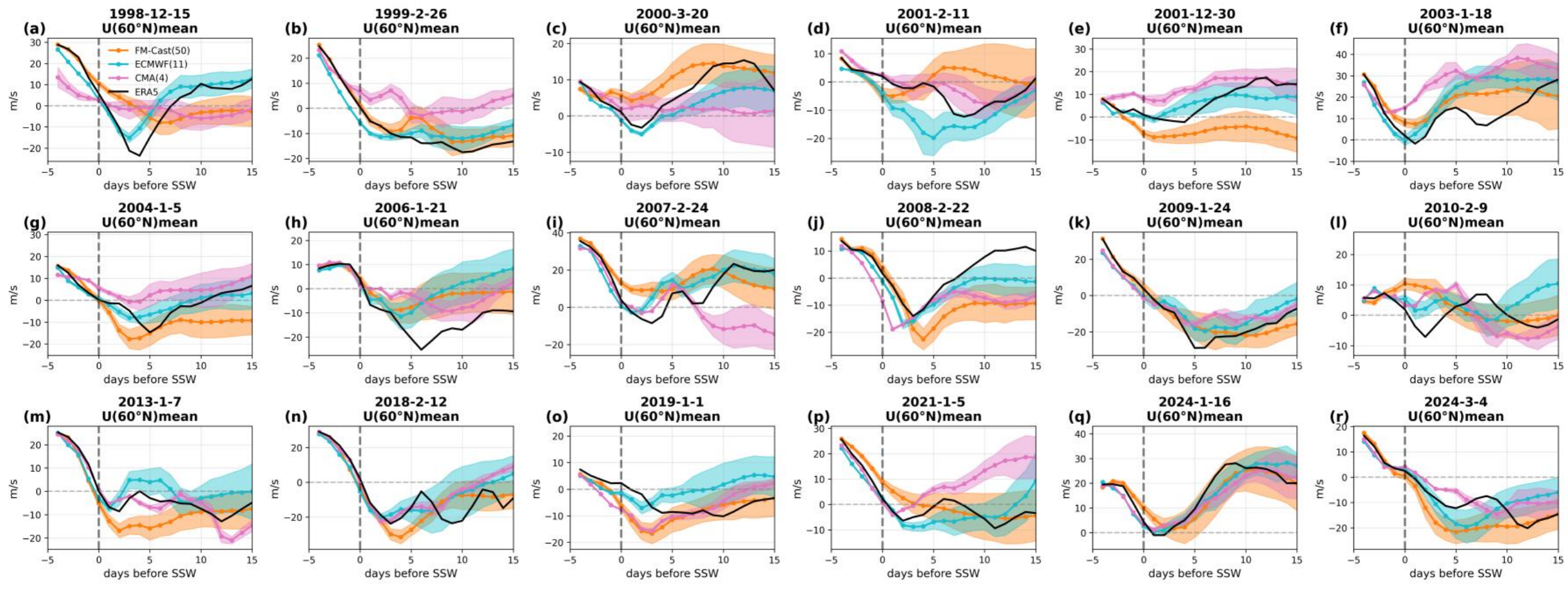

**Fig. S6 | Forecast skill comparison between FM-Cast and leading NWP models for the polar vortex index (10 hPa, 60°N zonal mean zonal wind) at a lead time of 5 days for 18 SSW events.** Same as Fig. S4.

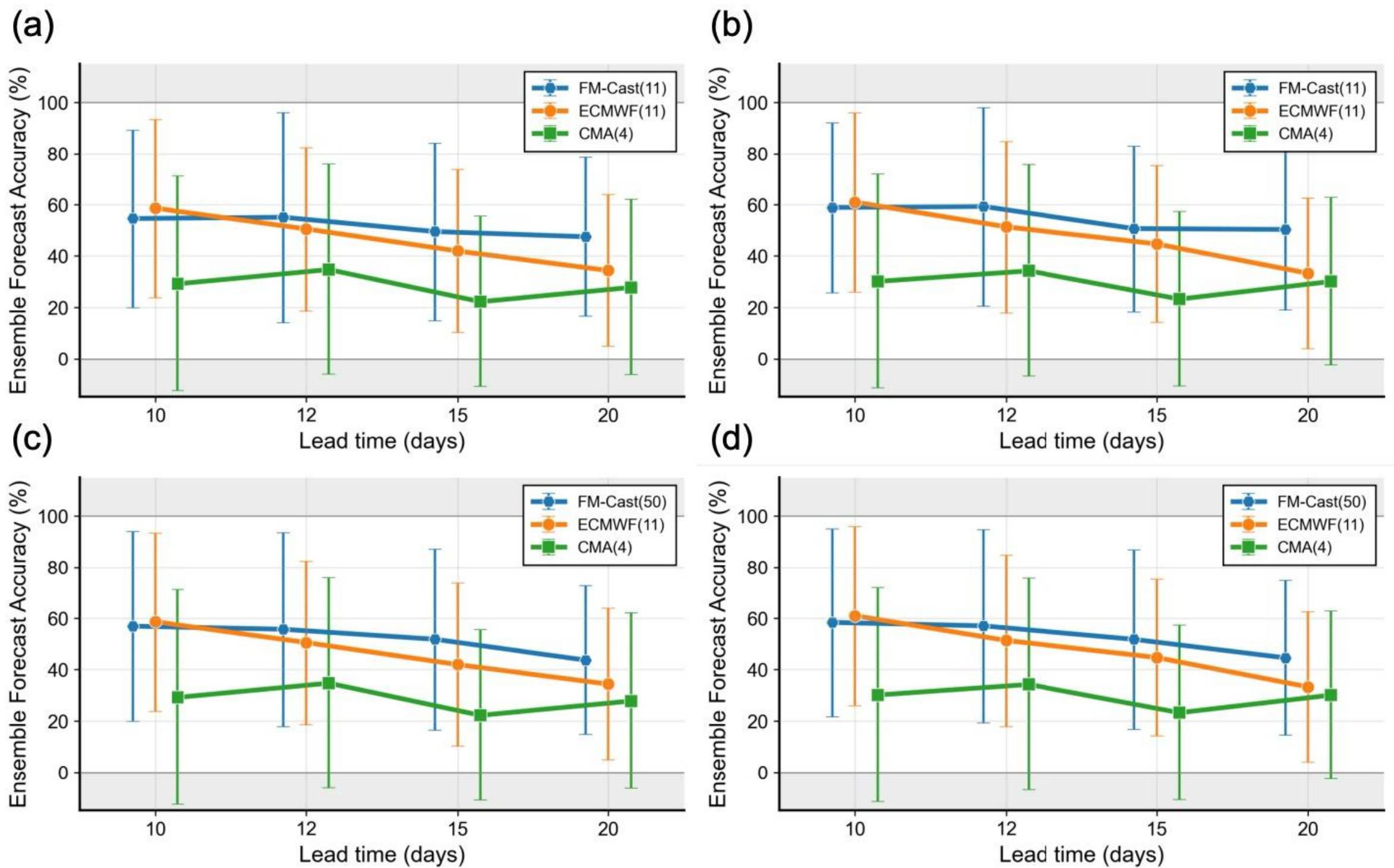

**Fig. S7 | Sensitivity analysis of ensemble forecast accuracy for 18 major SSW events comparing FM-Cast, ECMWF, and CMA under different ensemble sizes and probability calculation methods.** The y-axis represents the forecast accuracy (percentage of successful predictions), and the x-axis represents the forecast lead time (days). Error bars indicate the standard deviation across the 18 events. (a) Comparison using 11 ensemble members for both FM-Cast and ECMWF (4 for CMA), calculated using the Frequency Counting Method. (b) Comparison using 11 ensemble members for both FM-Cast and ECMWF (4 for CMA), calculated using the Gaussian PDF-based Method. (c) Comparison using 50 ensemble members for FM-Cast versus 11 for ECMWF (4 for CMA), calculated using the Frequency Counting Method. (d) Comparison using 50 ensemble members for FM-Cast versus 11 for ECMWF (4 for CMA), calculated using the Gaussian PDF-based Method. The consistent performance across all panels demonstrates that FM-Cast's forecast skill is robust and not primarily driven by its larger ensemble size relative to the NWP hindcasts.

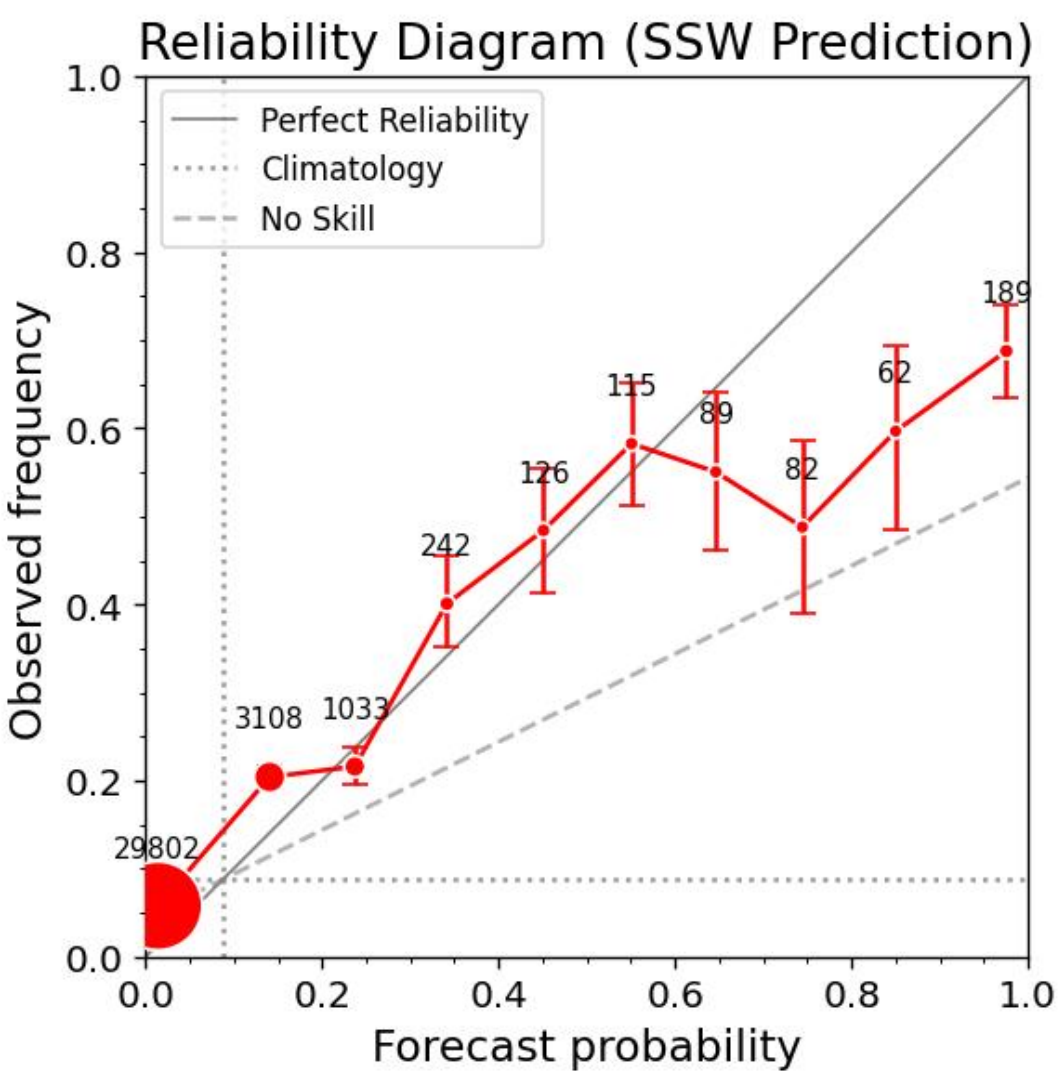

**Fig. S8 | Reliability diagram for FM-Cast SSW predictions over the period 1998–2024.** The x-axis represents the forecasted probability of an SSW occurrence, and the y-axis represents the observed relative frequency of the events. The diagonal gray solid line indicates perfect reliability (where forecast probability equals observed frequency). The red line traces the model's calibration curve, showing the correspondence between the model's confidence and the actual outcome.

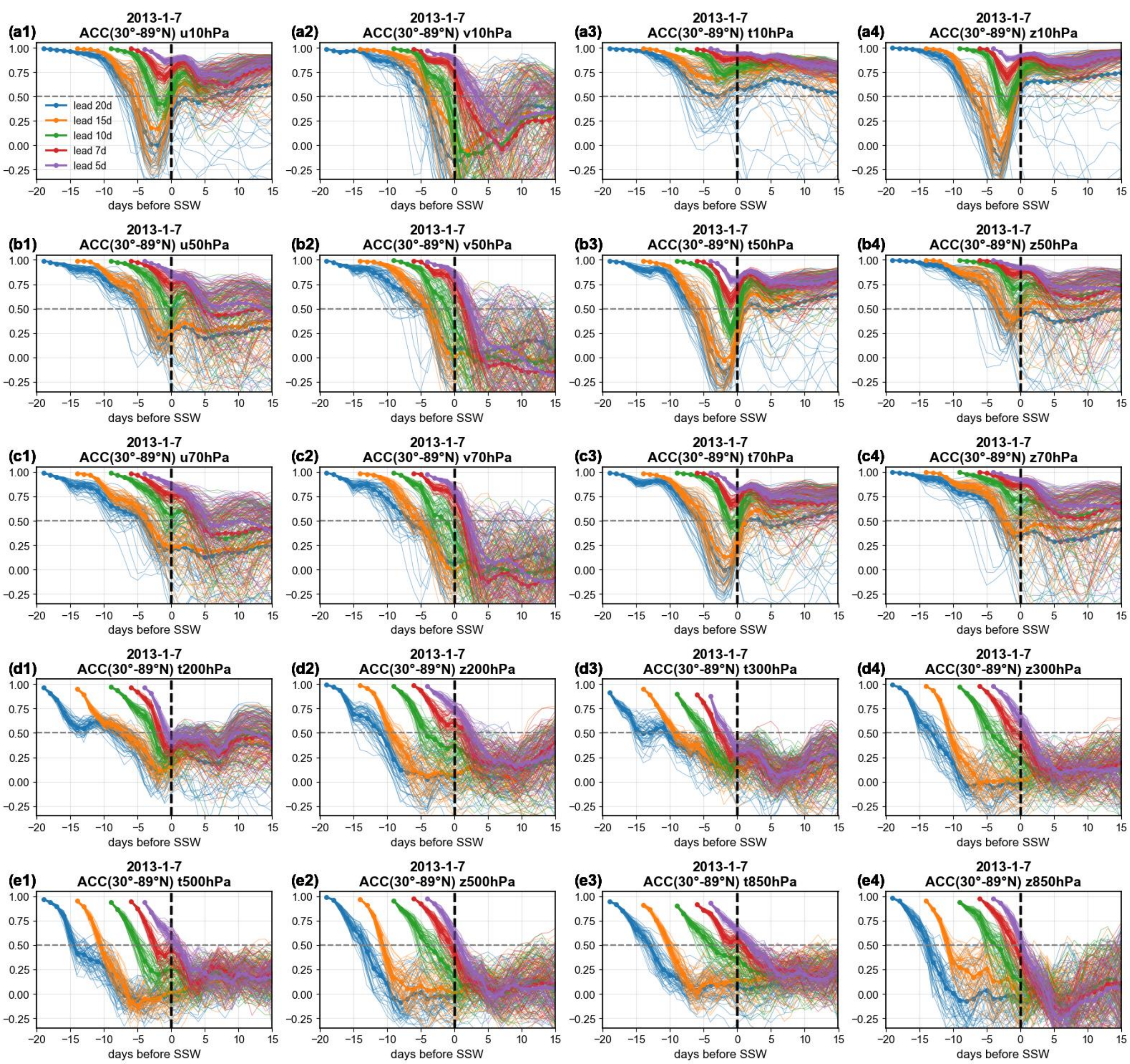

**Fig. S9 | Evolution of the Anomaly Correlation Coefficient (ACC) between FM-Cast forecasts and the ERA5 verification for variables at different levels for the SSW event on 7 January 2013.** The rows from top to bottom (a–g) correspond to the 10 hPa, 50 hPa, 70 hPa, 200 hPa (d1–d2) & 300 hPa (d3–d4), and 500 hPa (e1–e2) & 850 hPa (e3–e4) levels, respectively. The colored lines denote different forecast lead times (blue: 20 days, orange: 15 days, green: 10 days, red: 7 days, purple: 5 days); thin lines represent the 50 individual members, and thick dotted lines represent the ensemble mean. The x-axis shows the number of days relative to the SSW onset date (vertical dashed line). The horizontal gray dashed line indicates an ACC of 0.5.

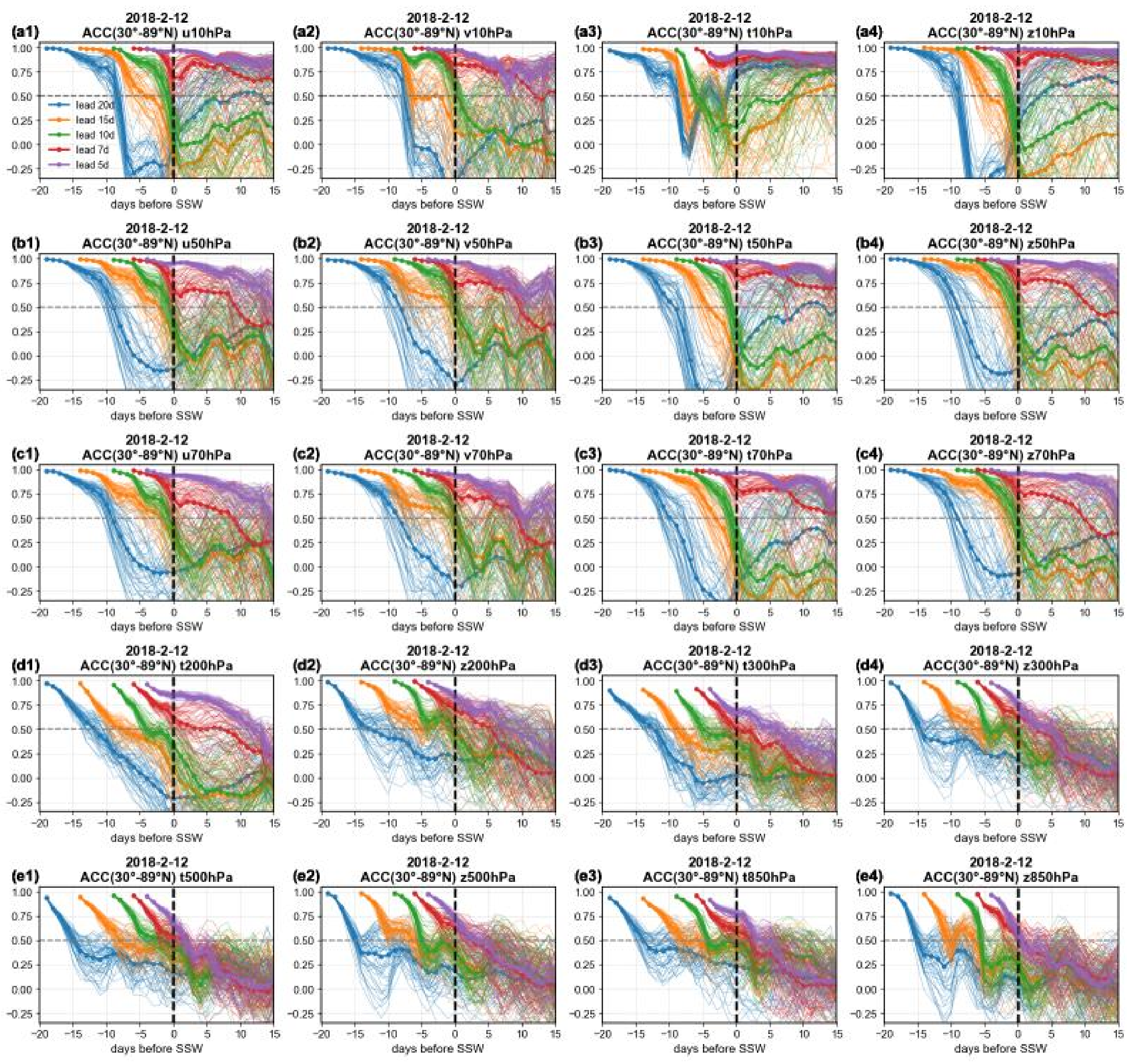

**Fig. S10 | Evolution of the Anomaly Correlation Coefficient (ACC) between FM-Cast forecasts and the ERA5 verification for variables at different levels for the SSW event on 12 February 2018.** Same as Fig. 10.

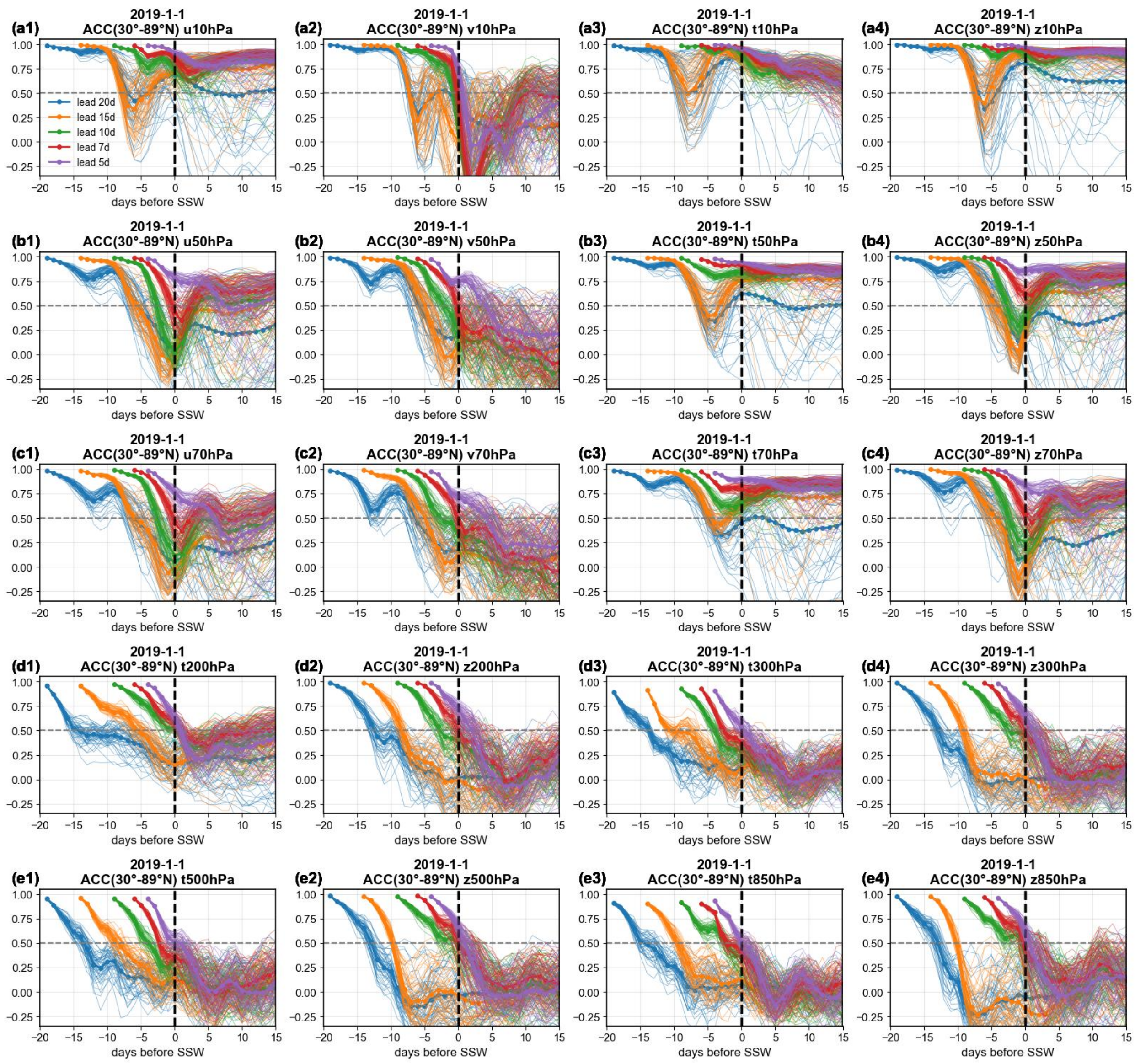

**Fig. S11 | Evolution of the Anomaly Correlation Coefficient (ACC) between FM-Cast forecasts and the ERA5 verification for variables at different levels for the SSW event on 1 January 2019.** Same as Fig. 10.

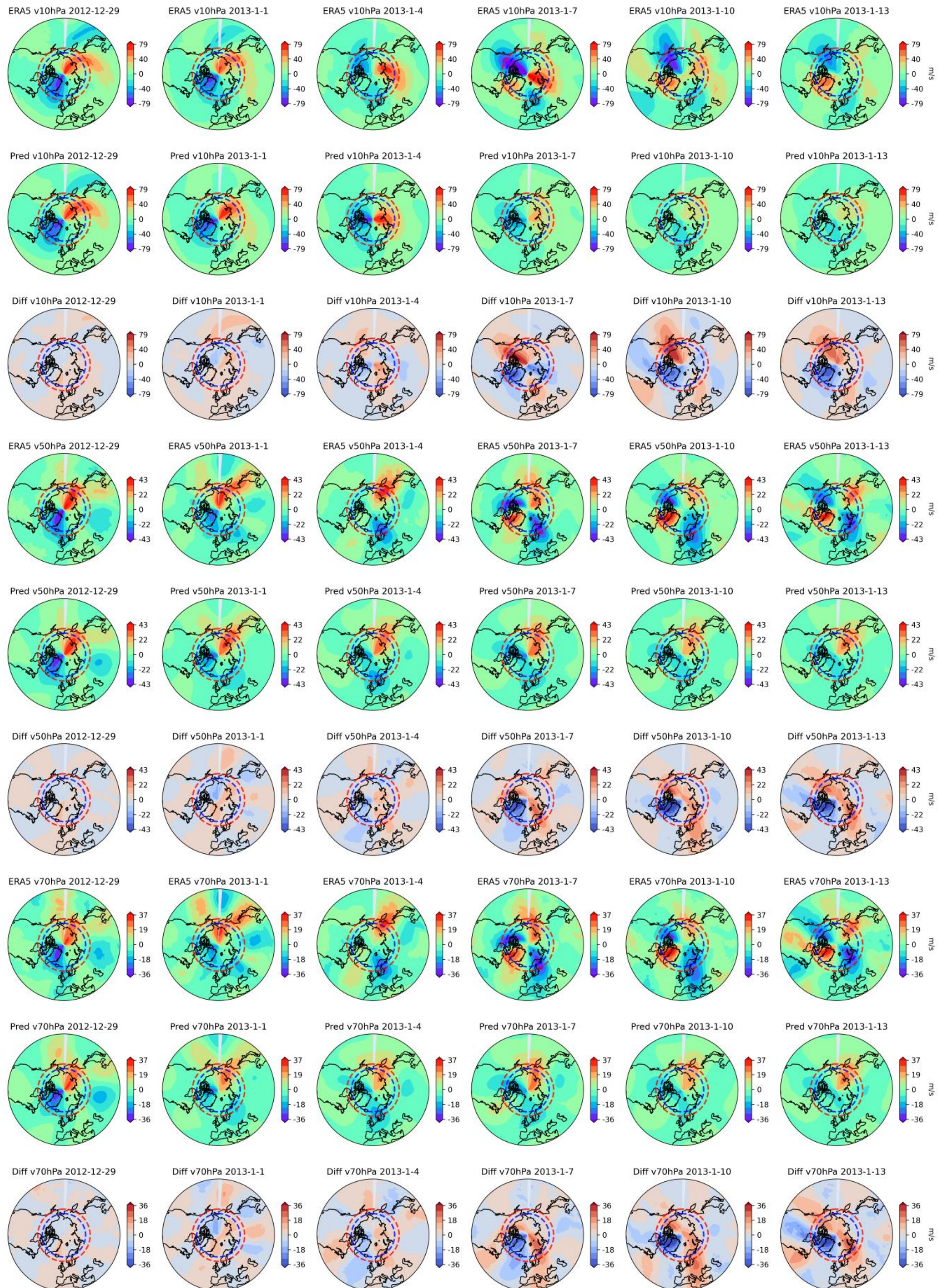

**Fig. S12 | Forecast performance of FM-Cast for the meridional wind field during a split-type SSW event (7 January 2013).** Shown are forecasts from 70 hPa to 10 hPa over the 30°N–89°N domain. The forecast was initialized on 25 and 26 December 2012. The columns from left to right are separated by 3-day intervals. The first, fourth, and seventh rows display the ERA5 field, while the second, fifth, and eighth rows display the corresponding forecasts (Pred.), the third, sixth, ninth rows display the difference between pred and ERA5, for the 10 hPa, 50 hPa, and 70 hPa levels, respectively. The results shown are the ensemble mean of the 50-member ensemble.

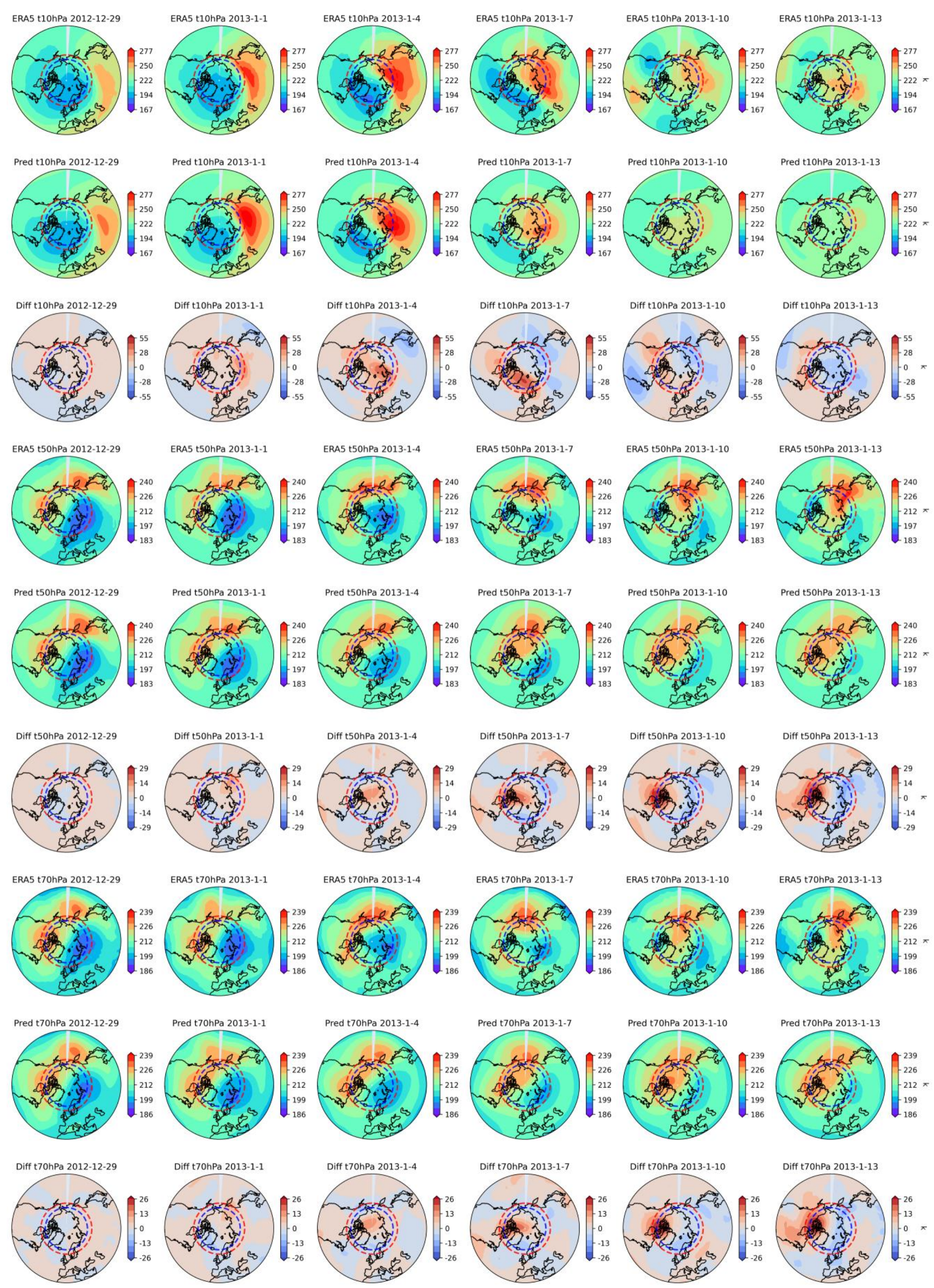

**Fig. S13 | Forecast performance of FM-Cast for the temperature field during a split-type SSW event (7 January 2013).** Same as Fig. S13.

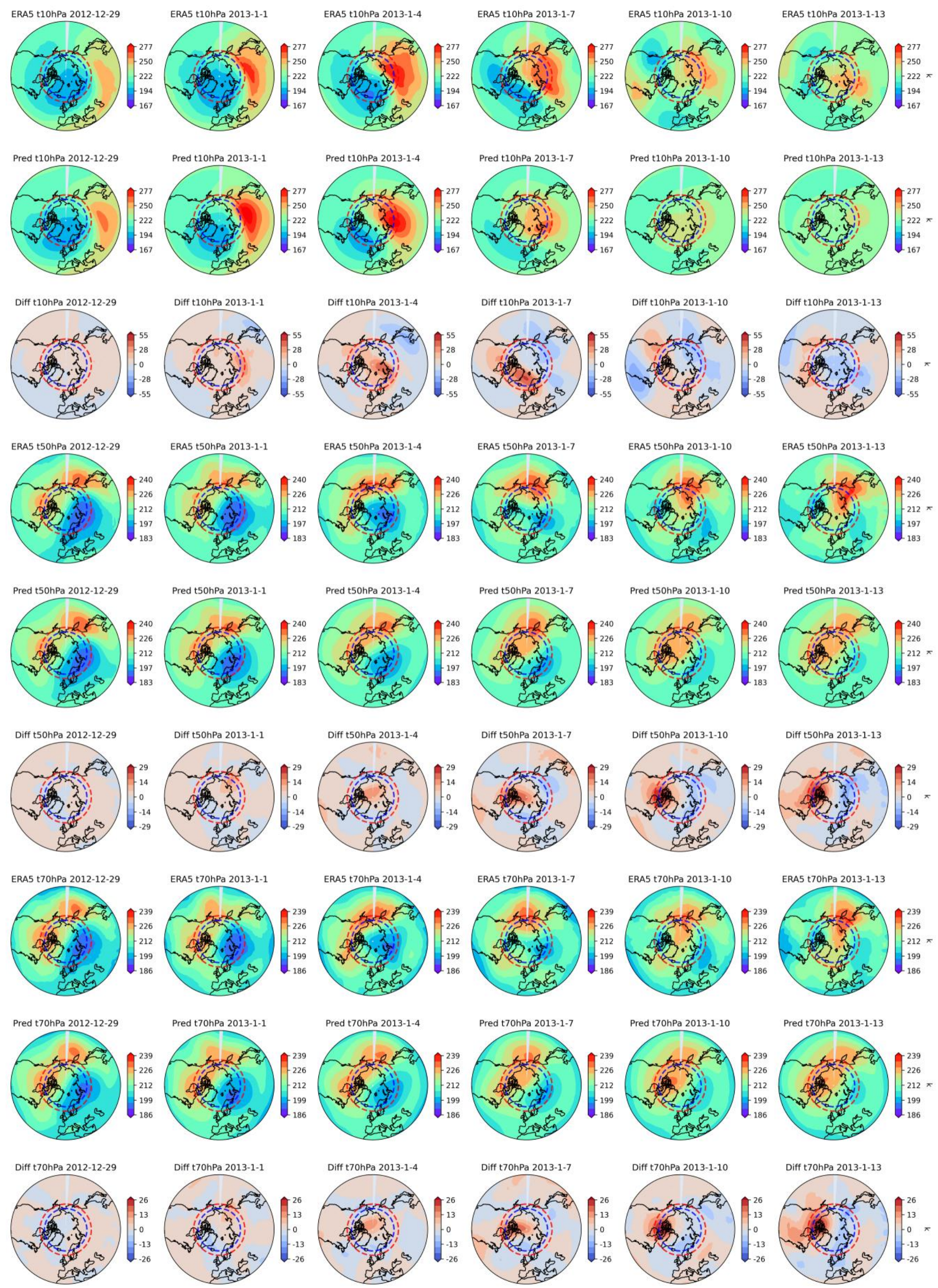

**Fig. S14 | Forecast performance of FM-Cast for the geopotential height field during a split-type SSW event (7 January 2013).** Same as Fig. S13.

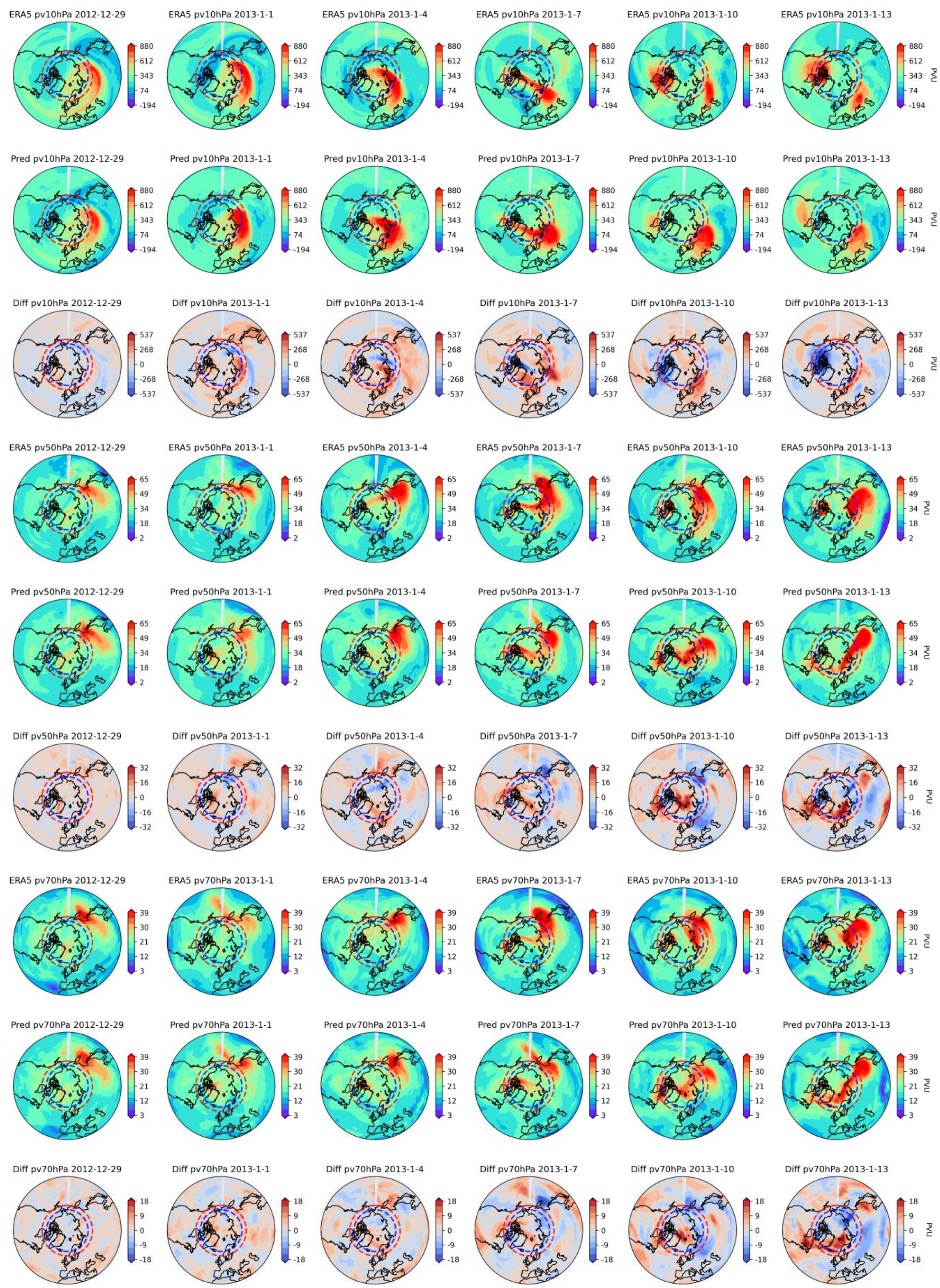

**Fig. S15 | Forecast performance of FM-Cast for the potential vorticity field during a split-type SSW event (7 January 2013).** Same as Fig. S13 but the results shown are the 30th member of the 50-ensemble member .

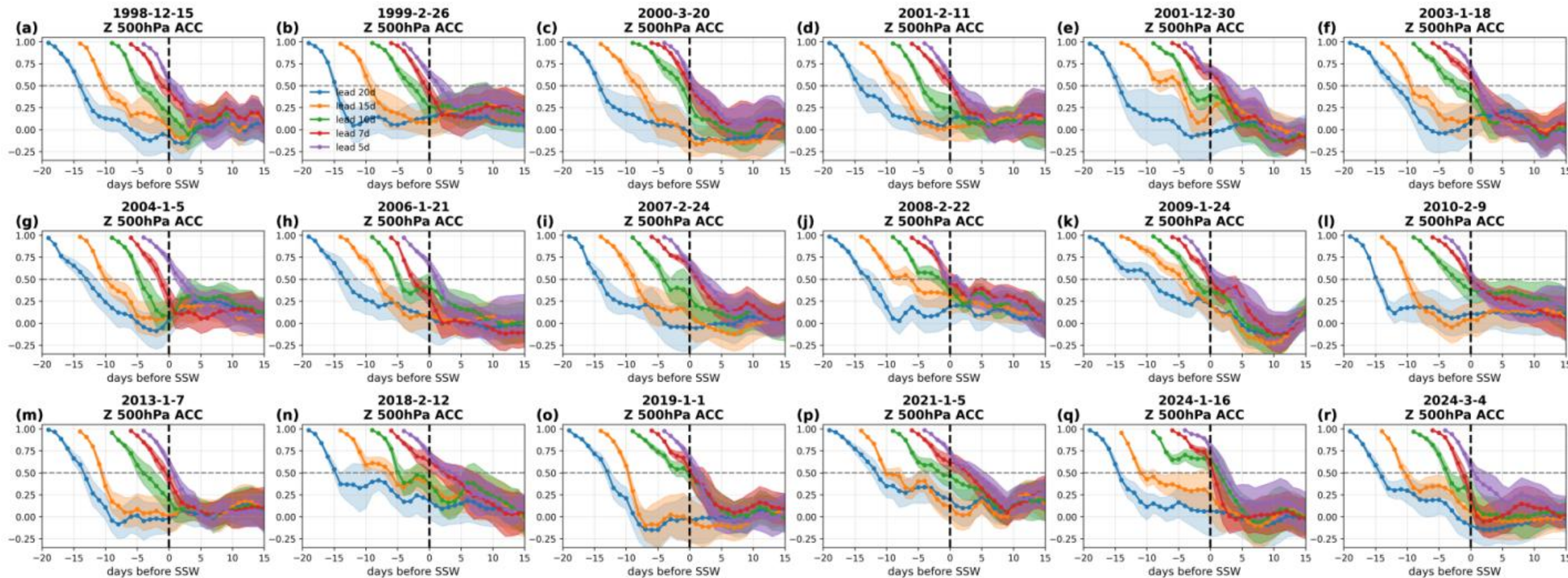

**Fig. S16 | Evolution of the Anomaly Correlation Coefficient (ACC) for 500 hPa geopotential height (Z500) during the 18 major SSW events from 1998–2024.** The x-axis represents the number of days relative to the SSW onset date (negative values denote days before onset, positive values denote days after, and the vertical black dashed line marks the onset day). The colored solid lines with markers represent the 50-member ensemble mean ACC for different forecast lead times (blue: 20 days, orange: 15 days, green: 10 days, red: 7 days, purple: 5 days), while the shaded areas indicate the ensemble standard deviation. The horizontal gray dashed line indicates the ACC threshold of 0.5, often used as a limit for useful prediction skill.

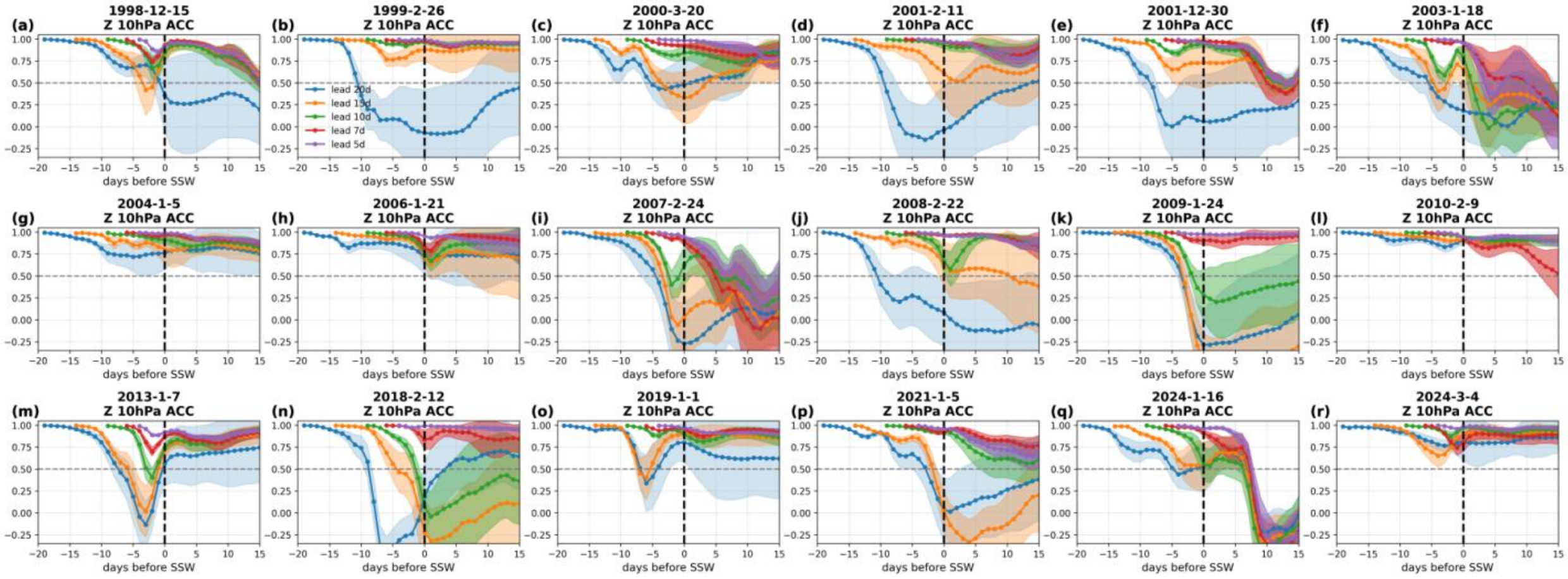

**Fig. S17 | Evolution of the Anomaly Correlation Coefficient (ACC) for 10 hPa geopotential height (Z10) during the 18 major SSW events from 1998–2024.** Same as Fig. S17.

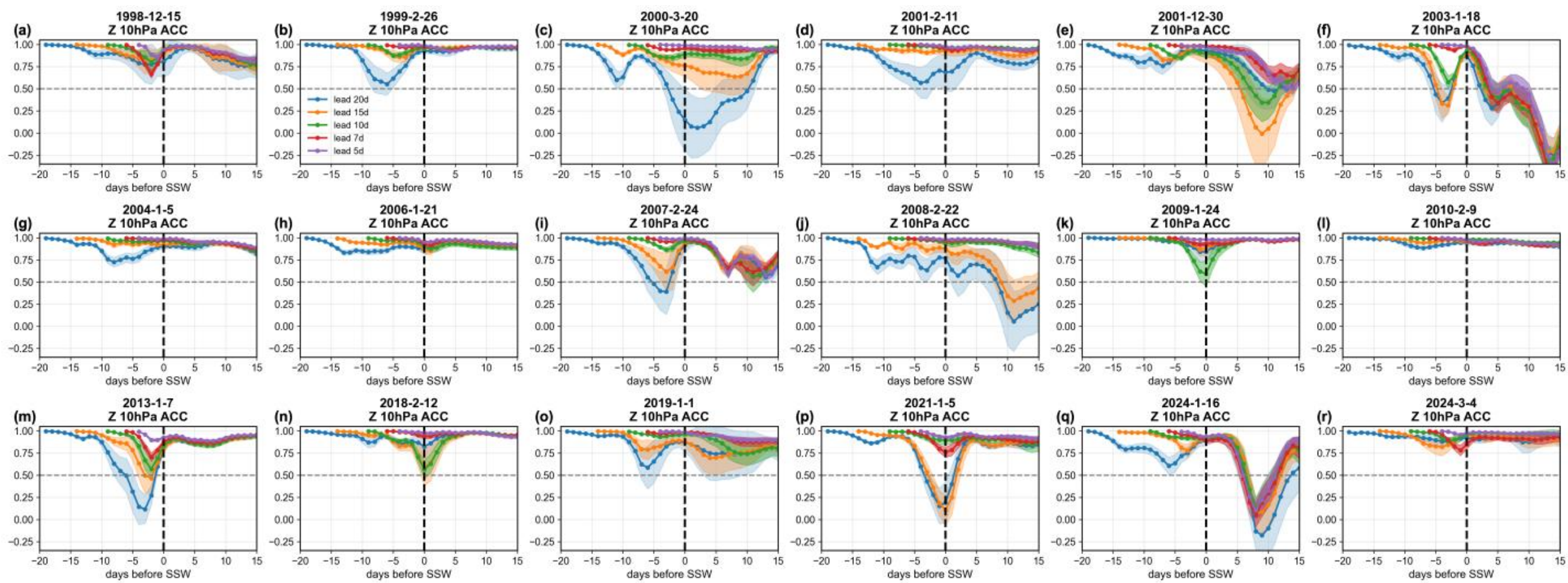

**Fig. S18 | Evolution of the Anomaly Correlation Coefficient (ACC) for 10 hPa geopotential height (Z10) during the 18 major SSW events from 1998–2024 under 'perfect troposphere' forcing.** Same as Fig. S17.

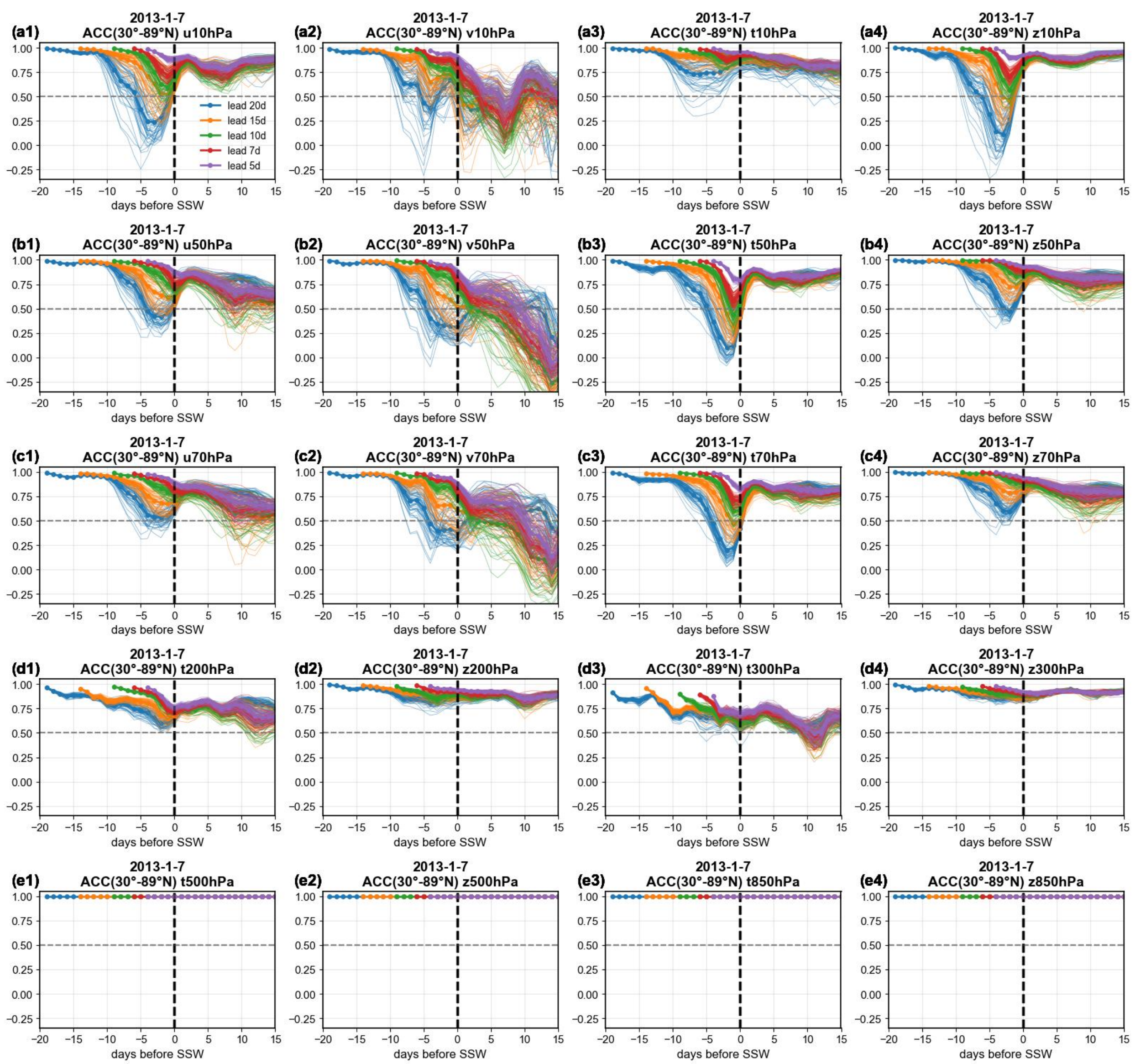

**Fig. S19 | Evolution of the Anomaly Correlation Coefficient (ACC) between FM-Cast forecasts and the ERA5 verification for variables at different levels for the SSW event on 7 January 2013 under 'perfect troposphere' forcing.** The rows from top to bottom (a–g) correspond to the 10 hPa, 50 hPa, 70 hPa, 200 hPa (d1–d2) & 300 hPa (d3–d4), and 500 hPa (e1–e2) & 850 hPa (e3–e4) levels, respectively. The colored lines denote different forecast lead times (blue: 20 days, orange: 15 days, green: 10 days, red: 7 days, purple: 5 days); thin lines represent the 50 individual members, and thick dotted lines represent the ensemble mean. The x-axis shows the number of days relative to the SSW onset date (vertical dashed line). The horizontal gray dashed line indicates an ACC of 0.5.

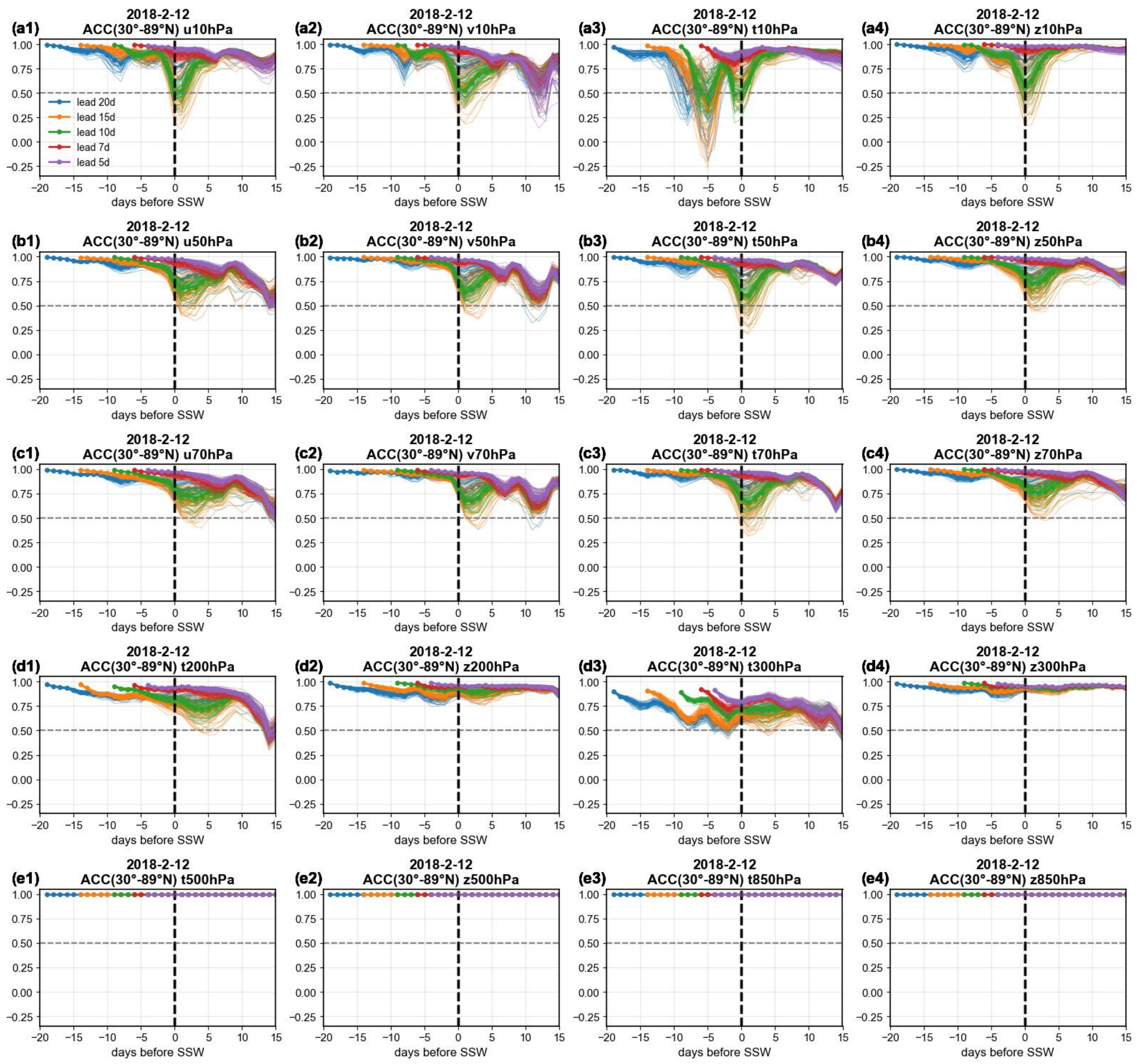

**Fig. S20 | Evolution of the Anomaly Correlation Coefficient (ACC) between FM-Cast forecasts and the ERA5 verification for variables at different levels for the SSW event on 12 February 2018 under 'perfect troposphere' forcing.** Same as Fig. S20.

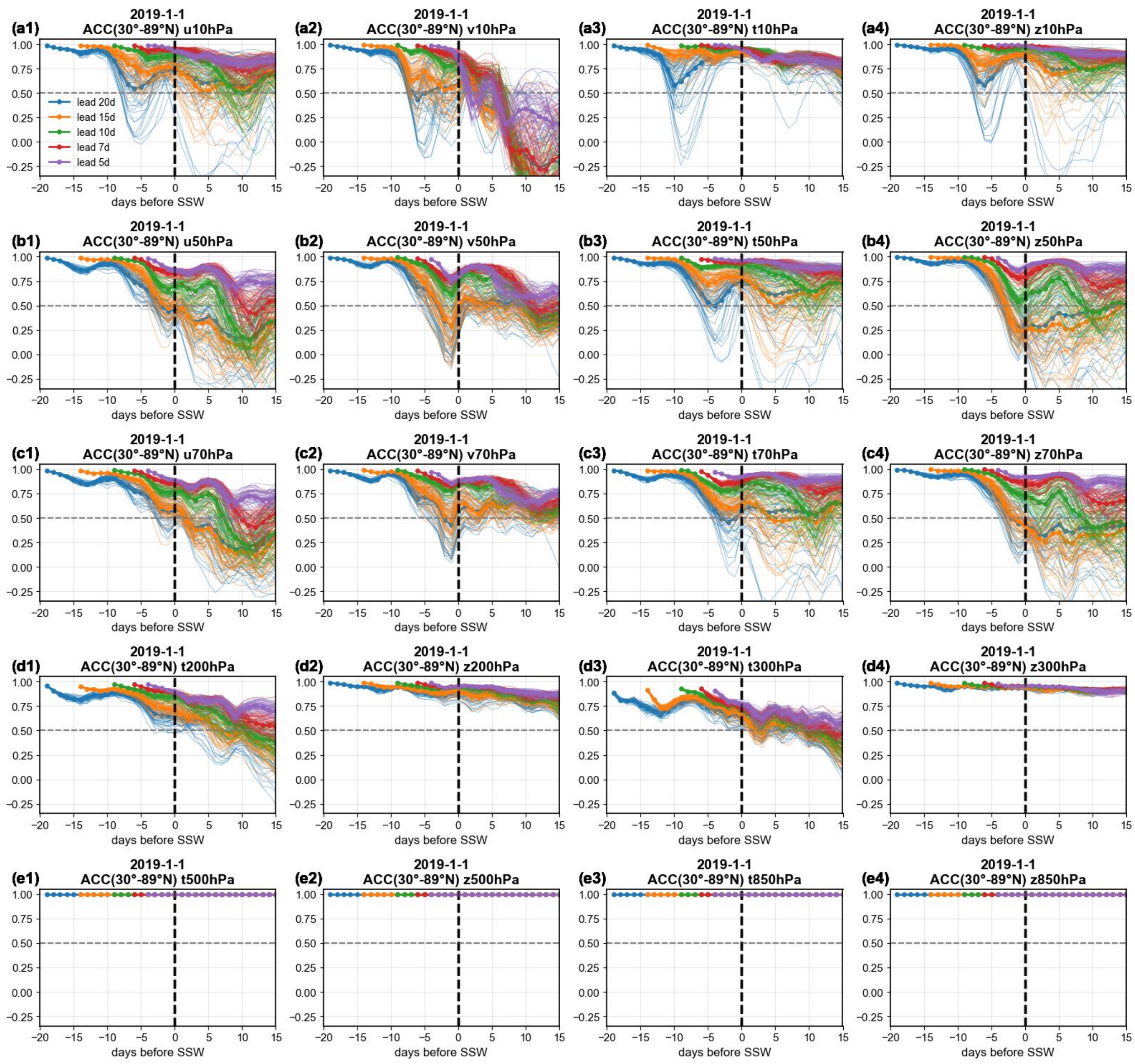

**Fig. S21 | Evolution of the Anomaly Correlation Coefficient (ACC) between FM-Cast forecasts and the ERA5 verification for variables at different levels for the SSW event on 1 January 2019 under 'perfect troposphere' forcing.** Same as Fig. S20.

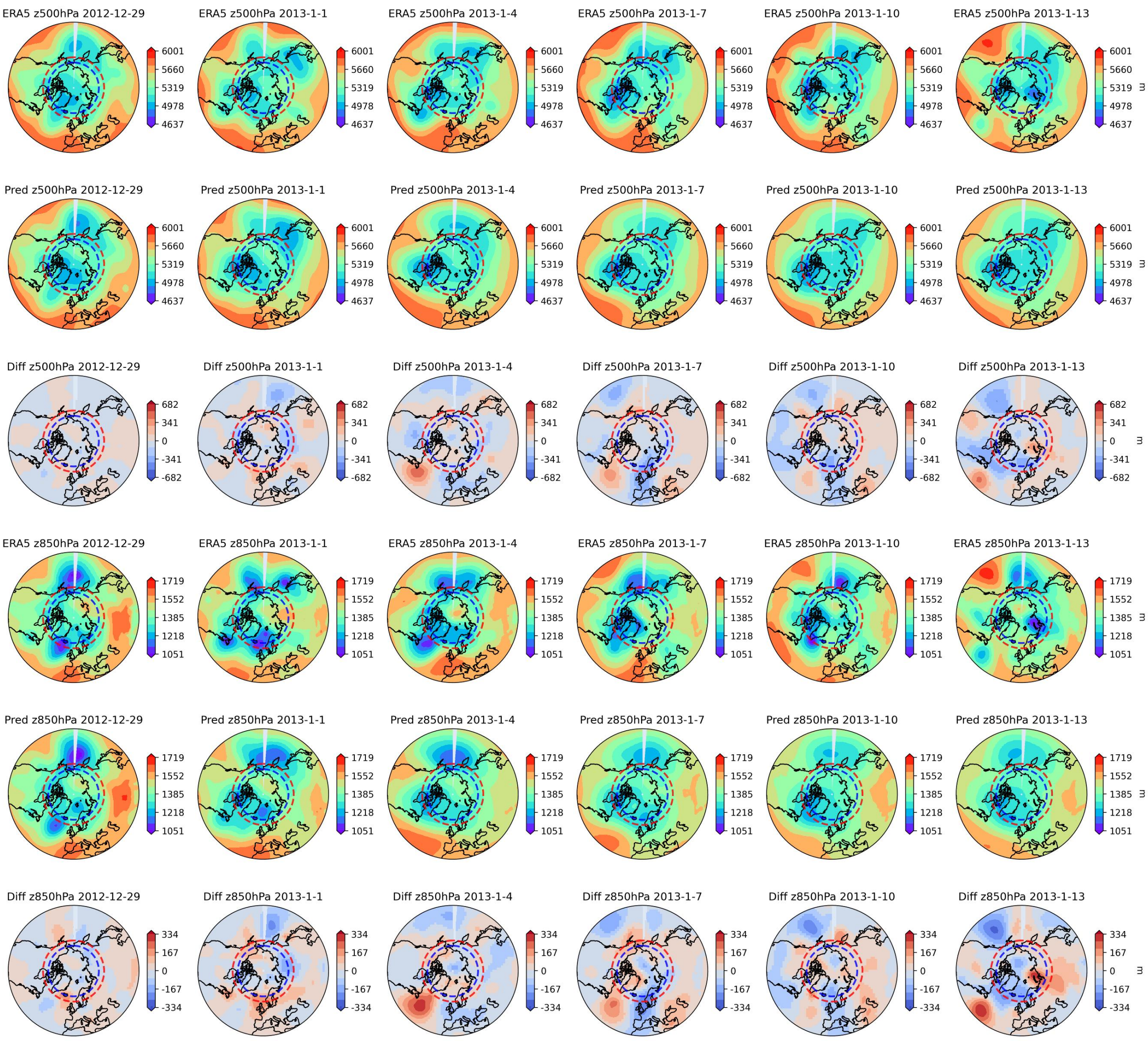

**Fig. S22 | Forecast performance of FM-Cast for the geopotential height field during a split-type SSW event (7 January 2013).** Shown are forecasts from 850 hPa to 500 hPa over the 30°N–89°N domain. The forecast was initialized on 25 and 26 December 2012. The columns from left to right are separated by 3-day intervals. The first, fourth rows display the ERA5 field, while the second, fifth, rows display the corresponding forecasts (Pred.), the third, sixth rows display the difference between pred and ERA5, for the 500 hPa, 850 hPa levels, respectively. The results shown are the ensemble mean of the 50-member ensemble.

# B. Supplementary Text

### Impact of Lower Stratospheric Cold Bias in ERA5 Reanalysis on Forecast Performance

The ERA5 reanalysis is known to exhibit a significant cold bias in the lower stratosphere during the 2000–2006 period, primarily due to the assimilation of specific satellite radiances (Simmons et al., 2020) [1]. Recognizing that data quality is paramount for data-driven deep learning models, we conducted a rigorous sensitivity experiment to quantify the potential impact of this bias on FM-Cast's forecast skill.

Experimental Design: We utilized the ERA5.1 reanalysis, a corrected version of ERA5 produced by ECMWF specifically to address the stratospheric temperature bias [1]. In this sensitivity experiment, we replaced the training and validation data for the affected period (2000–2006) with ERA5.1, while retaining standard ERA5 data for the remaining years (1979–1999 and 2007–2024). We then retrained the FM-Cast model from scratch using this "bias-corrected" dataset and re-evaluated the forecasts for all 18 major SSW events (1998–2024).

The results of this sensitivity analysis are presented in Figs. S23–S28. Fig. S23 displays the overall forecast performance. Figs. S24–S28 provide a detailed side-by-side comparison of the polar vortex index forecasts at lead times of 5, 7, 10, 15, and 20 days.

The comparison reveals that the choice of training data does impact the forecast outcomes. While the ensemble spread and general trends overlap for many events, distinct deviations are observed in specific cases. Notably, the model trained with the ERA5.1-corrected dataset tends to predict a stronger polar vortex (or a weaker/delayed breakdown) compared to the baseline model trained on standard ERA5 (e.g., see Fig. S25 k, q; Fig. S26 a, b, o; Fig. S27 a, b, e, m). The introduction of ERA5.1 data did not yield a systematic improvement in forecast skill metrics across the full test set, nor did it resolve the difficulties in predicting "marginal SSW" events.

We attribute the observed differences and the lack of systematic improvement to two primary factors:

Data Consistency: The ERA5.1 correction is applied only to the 2000–2006 period. Mixing ERA5.1 for this specific window with standard ERA5 data for the remaining years (1979–1999, 2007–2024) may introduce temporal inconsistencies or "jumps" in the data distribution. Deep learning models are highly sensitive to such inhomogeneities, which might offset the benefits of the bias correction in training data.

Sample Size Sensitivity: As discussed in the main text (see methodology section regarding cross-validation), the number of SSW samples available for training is extremely limited (21–23 events). Consequently, the model training is sensitive to even minor perturbations in the dataset. The shift in data distribution caused by introducing ERA5.1, while physically "correct" for the lower stratosphere, may have inadvertently introduced noise or distributional shifts into the learning process given the small sample size.

In summary, while the ERA5.1 dataset corrects known lower-stratospheric biases, its use in a hybrid training configuration leads to divergent forecast trajectories for certain SSW events without systematically enhancing predictive skill at the 10 hPa level. This sensitivity analysis objectively documents these variations, highlighting the profound influence of training data homogeneity and sample size constraints on data-driven stratospheric forecasting.

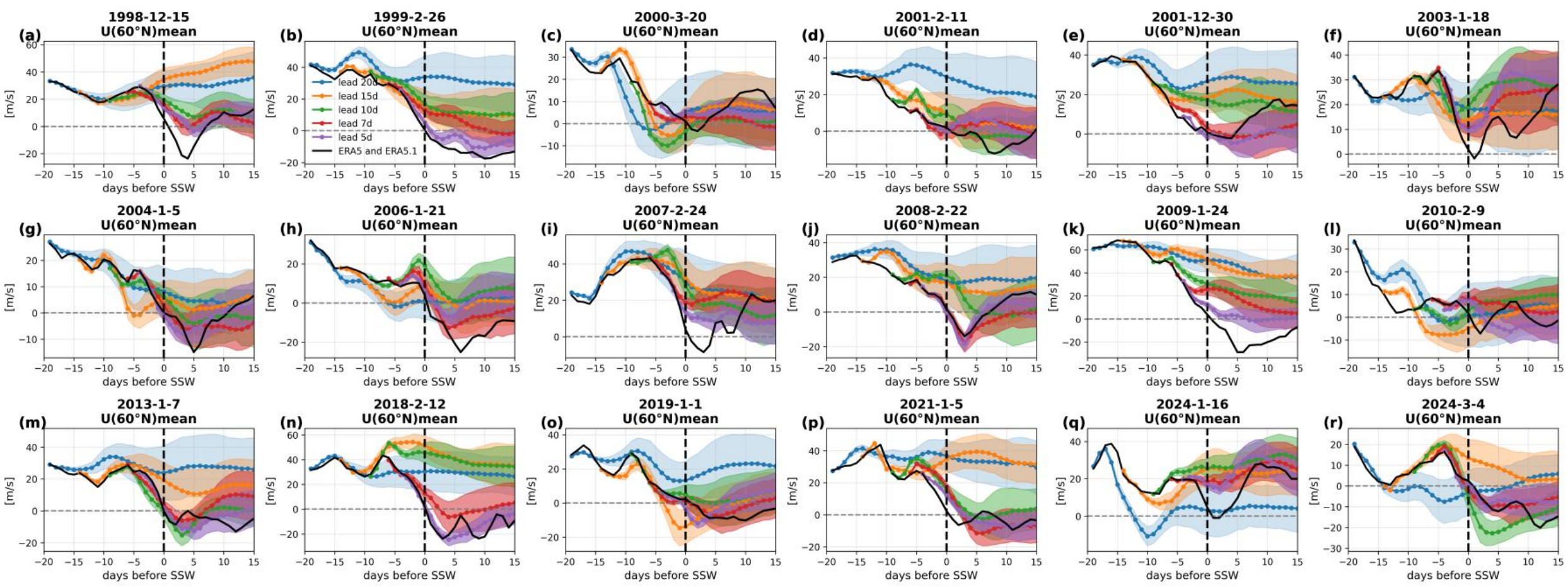

**Fig. S23 | FM-Cast forecasts of the polar vortex index (10 hPa, 60°N zonal-mean zonal wind) for the 18 major SSW events from 1998–2024 with the ERA5 verification (black solid line), trained using the bias-corrected ERA5.1 dataset for the 2000–2006 period.** The x-axis represents the number of days relative to the SSW onset date (negative values are days before onset, positive values are days after, and the vertical dashed line marks the onset day). The colored dotted lines represent the 50-member ensemble mean, while the shaded areas indicate the ensemble standard deviation. Different colors correspond to different forecast lead times (blue: 20 days, orange: 15 days, green: 10 days, red: 7 days, purple: 5 days). The horizontal gray dashed line indicates the zero-wind line.

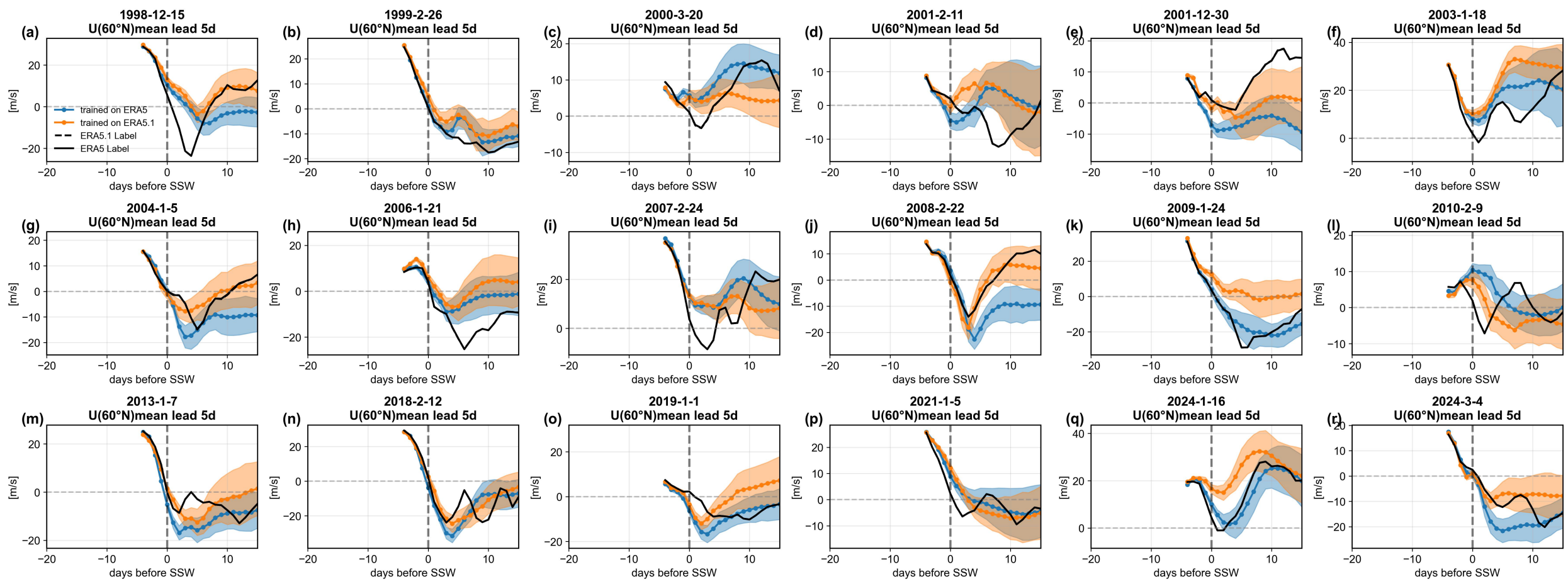

**Fig. S24 | Comparison of FM-Cast forecasts for the polar vortex index (10 hPa, 60°N zonal-mean zonal wind) at a 5-day lead time under different training data configurations.** The blue line with shading represents the ensemble mean forecast from the model trained on the standard ERA5 dataset. The orange line with shading represents the results from the model trained on the ERA5 dataset integrated with ERA5.1 (specifically correcting the stratospheric bias in the 2000–2006 period). The solid black line indicates the ground truth from ERA5, while the dashed black line indicates the ground truth from ERA5.1. Note that these two lines overlap almost completely and are visually indistinguishable, indicating that despite the cold anomalies in the lower stratosphere of ERA5 during this period, the differences at the upper level (10 hPa) are not significant. The shaded areas represent the ensemble standard deviation. The vertical gray dashed line marks the SSW onset date (day 0), and the horizontal gray dashed line indicates the zero-wind threshold.

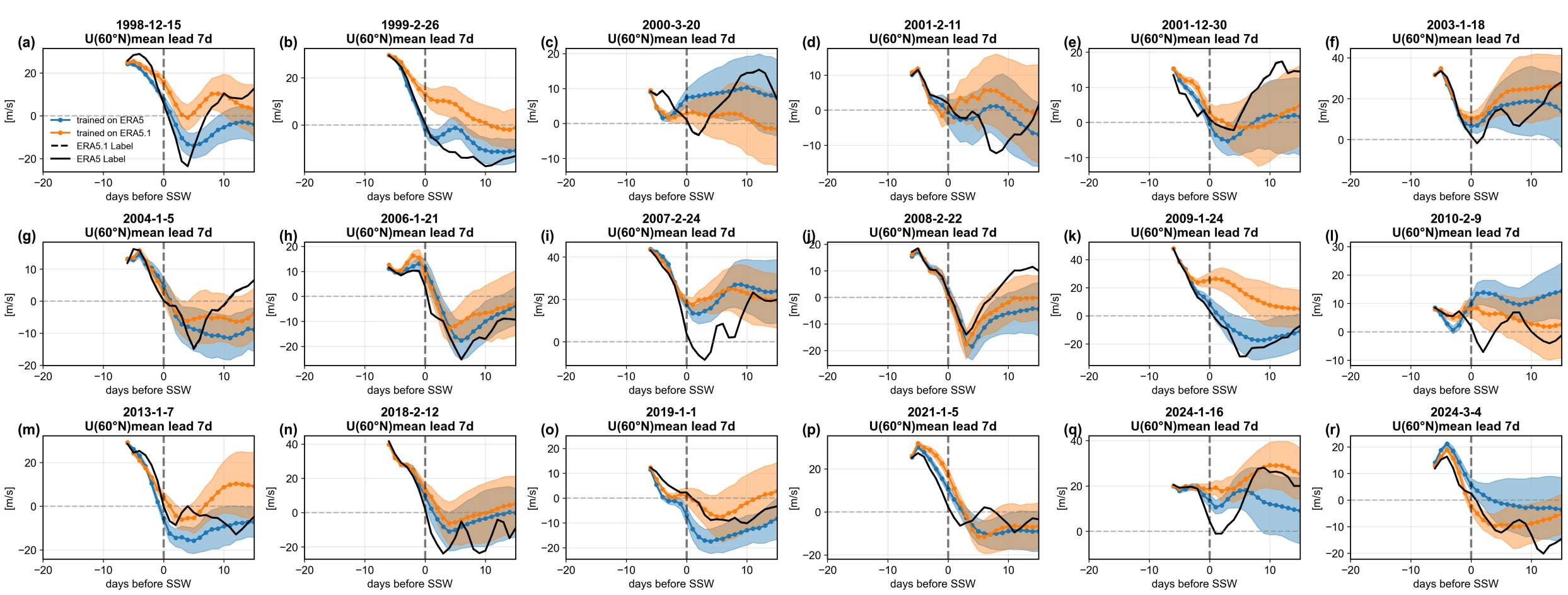

**Fig. S25 | Comparison of FM-Cast forecasts for the polar vortex index (10 hPa, 60°N zonal-mean zonal wind) at a 7-day lead time under different training data configurations.** Same as Fig. S24.

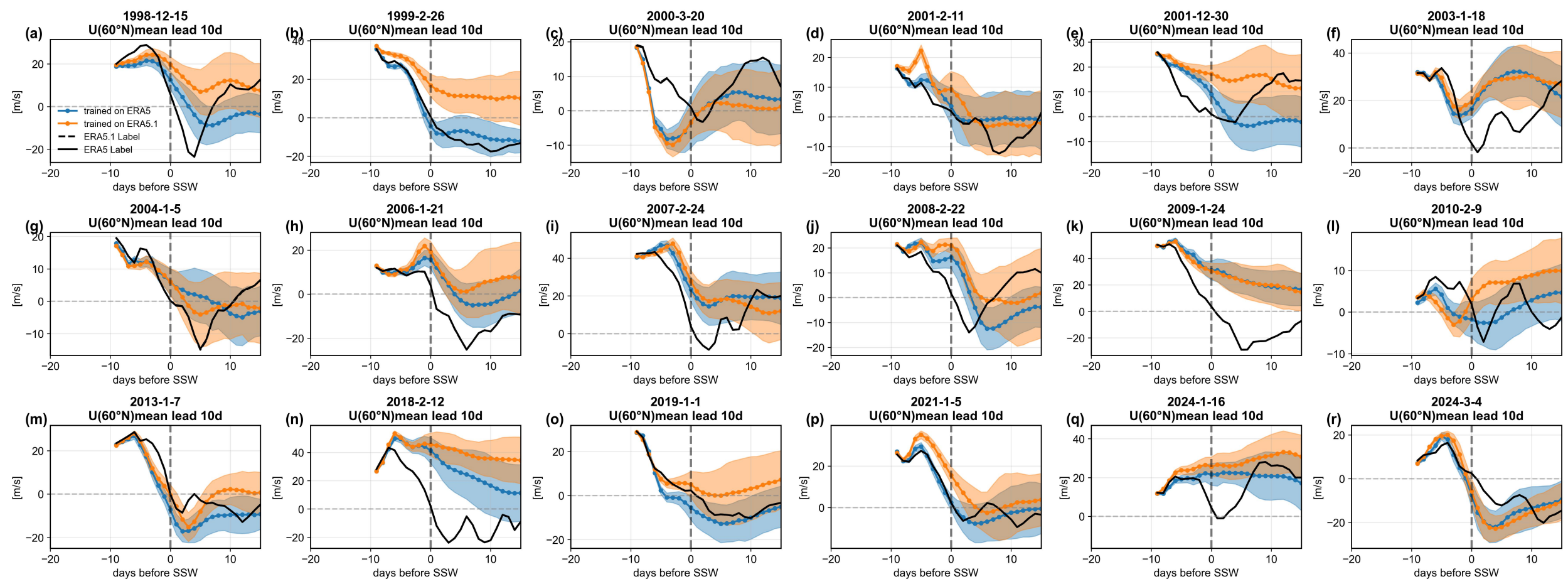

**Fig. S26 | Comparison of FM-Cast forecasts for the polar vortex index (10 hPa, 60°N zonal-mean zonal wind) at a 10-day lead time under different training data configurations.** Same as Fig. S24.

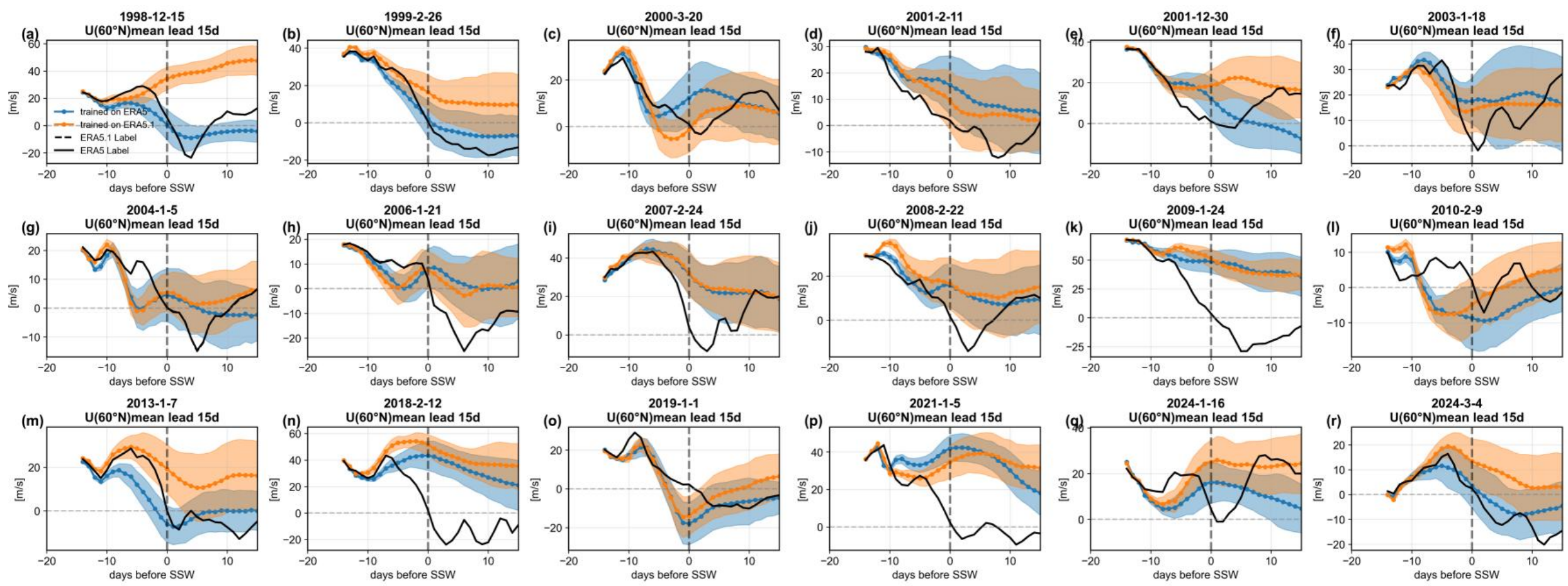

**Fig. S27 | Comparison of FM-Cast forecasts for the polar vortex index (10 hPa, 60°N zonal-mean zonal wind) at a 15-day lead time under different training data configurations.** Same as Fig. S24.

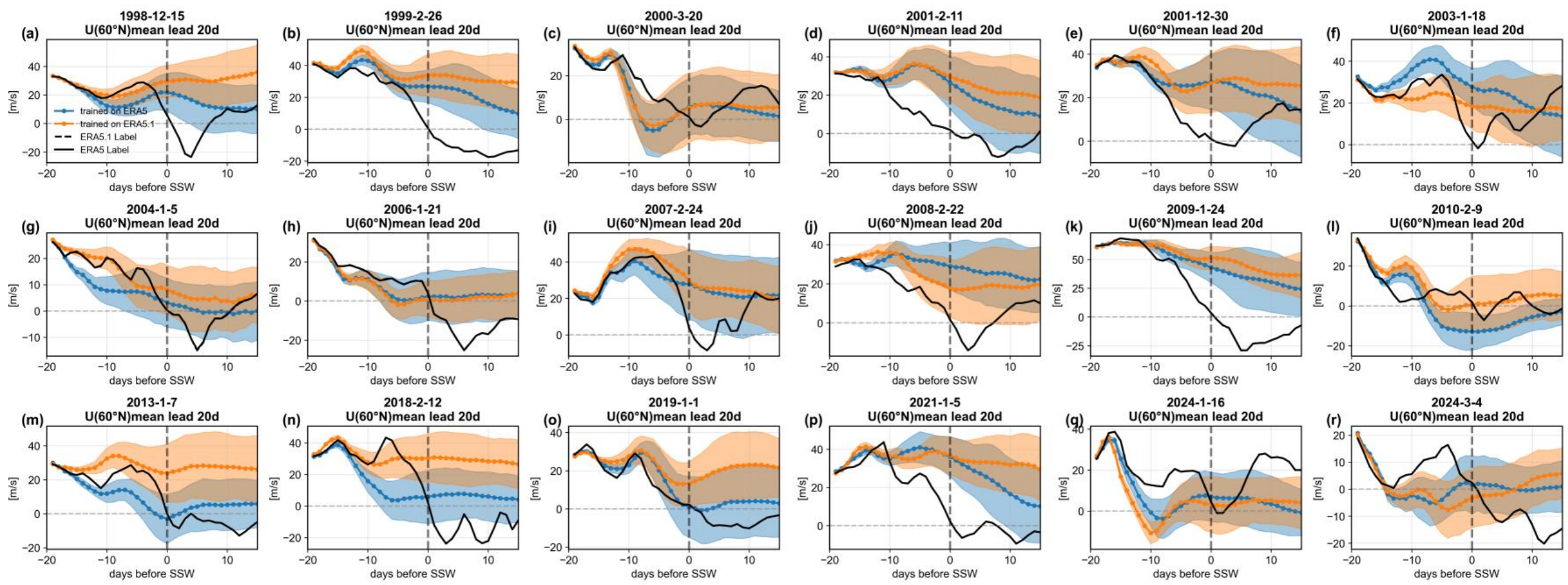

**Fig. S28 | Comparison of FM-Cast forecasts for the polar vortex index (10 hPa, 60°N zonal-mean zonal wind) at a 20-day lead time under different training data configurations.** Same as Fig. S24.

## Reference:

1, Simmons, A., Soci, C., Nicolas, J., Bell, B., Berrisford, P., Dragani, R., ... & Schepers, D. (2020).

Global stratospheric temperature bias and other stratospheric aspects of ERA5 and ERA5. 1.